\documentclass[review]{elsarticle}

\usepackage{lineno}
\usepackage{hyperref}       
\usepackage[utf8]{inputenc} 
\usepackage{url}            
\usepackage{amsfonts}       
\usepackage{nicefrac}       
\usepackage{microtype}      
\usepackage{lipsum}
\usepackage{fancyhdr}       
\usepackage{graphicx}       
\usepackage{subfigure}      
\usepackage{amsmath}        
\usepackage{amssymb}        
\usepackage{enumerate}      
\usepackage{geometry}       
\usepackage{booktabs}       
\usepackage{color}          
\usepackage{multirow}       
\usepackage{threeparttable}  
\usepackage{color, colortbl} 
\usepackage{rotating}
\usepackage[dvipsnames]{xcolor}
\usepackage{makecell}
\modulolinenumbers[5]

\definecolor{Gray}{gray}{0.9}
\definecolor{LightCyan}{rgb}{0.88,1,1}

\hypersetup{pdfauthor=author}

\journal{Journal of Cell Patterns}

\usepackage{numcompress}
\begin{document}

\begin{frontmatter}

\title{Are we ready for a new paradigm shift? \\ A Survey on Visual Deep MLP}

\author[mymainaddress]{Ruiyang Liu\corref{myequalauthor}}
\author[mysecondaryaddress]{Yinghui Li\corref{myequalauthor}}
\author[mymainaddress]{Linmi Tao\fnref{mycorrespondingauthor}}
\author[mymainaddress]{Dun Liang} 
\author[mysecondaryaddress]{Hai-Tao Zheng}

\address[mymainaddress]{Department of Computer Science and Technology, BNRist, Tsinghua University \& \\ Key Lab of Pervasive Computing, Ministry of Education of China, \\Beijing {\rm 100084}, China}
\address[mysecondaryaddress]{Tsinghua Shenzhen International Graduate School, Tsinghua University, \\ Shenzhen {\rm 518055}, China}

\cortext[myequalauthor]{Equal contribution: \{lry20, liyinghu20\}@mails.tsinghua.edu.cn }
\fntext[mycorrespondingauthor]{Corresponding author: linmi@tsinghua.edu.cn}





\begin{abstract}
Recently, the proposed deep MLP models have stirred up a lot of interest in the vision community. Historically, the availability of larger datasets combined with increased computing capacity leads to paradigm shifts. This review paper provides detailed discussions on whether MLP can be a new paradigm for computer vision. We compare the intrinsic connections and differences between convolution, self-attention mechanism, and Token-mixing MLP in detail. Advantages and limitations of Token-mixing MLP are provided, followed by careful analysis of recent MLP-like variants, from module design to network architecture, and their applications. In the GPU era, the locally and globally weighted summations are the current mainstreams, represented by the convolution and self-attention mechanism, as well as MLP. We suggest the further development of paradigm to be considered alongside the next-generation computing devices.
\end{abstract}


\end{frontmatter}


\section{Introduction}

In computer vision, the ambition to create a system that imitates how the brain perceives and understands visual information fueled the initial development of neural networks~\cite{mcculloch1943logical,rosenblatt1958perceptron}. Subsequently, Convolutional Neural Networks (CNNs)~\cite{fukushima1979neural,lecun1989backpropagation,lecun1998gradient}, Multilayer Perceptrons (MLPs)~\cite{rumelhart1985learning} and Boltzmann Machine~\cite{ackley1985learning,smolensky1986information} were proposed, and achieved fruitful results in theoretical researches~\cite{cybenko1989approximation,hornik1989multilayer,funahashi1989approximate,hinton1986learning,DBLP:journals/neco/Hinton89,tanaka1998mean} in the last century. CNNs stood out due to their computational efficiency over MLPs and Deep Boltzmann Machines in the contest to replace handcrafted features, and topped the list for vast visual tasks in the 2010s. From 2020, the Transformer-based models introduced from the natural language processing field to the visual field have once again reached a new peak. With the introduction of MLP-Mixer~\cite{tolstikhin2021mlp} in 2021, the hot topic in the vision community comes: \emph{Will MLP become a new paradigm and push computer vision to a new height}? This survey aims to provide opinions on this issue.

From a historical perspective, the availability of larger datasets combined with the transition from CPU-based training to GPU-based training leads to paradigm shifts and a gradual reduction in human intervention. The locally weighted summation represented by convolution and globally weighted summation represented by self-attention are the current mainstreams. The Token-mixing MLP~\cite{tolstikhin2021mlp} in MLP-Mixer further abandons the artificially designed self-attention mechanism and lets the model learn the global weights matrix autonomously from the raw data, seemly in line with the laws of historical development. 

We review MLP-Mixer in detail, compare the intrinsic connections and differences between convolution, self-attention mechanism, and Token-mixing MLP. We observe that the Token-mixing MLP is an enlarged and weights-shared-between-channel version of depthwise convolution~\cite{chollet2017xception}, which faces challenges such as high computational complexity and resolution sensitivity. Exhaustive analysis reveals that, not only the recent MLP-like variant designs are gradually approaching the direction of CNN, but the performance of these variants in visual tasks still lags behind CNN and Transformer-based. At this moment, MLP is not a new paradigm that can push computer vision to new heights. In fact, computing paradigm and computing hardware are cooperative. The current weighted-sum paradigms have driven the booming of GPU-based computing and deep learning itself, while we believe the next paradigm or Boltzmann-like will also grow up with a new generation of computing hardware.

The rest of the paper is organized as follows. Section~\ref{sec:preliminary} reviews MLP, CNN, and Transformer, as well as their corresponding paradigms from a historical perspective. Section~\ref{sec:Pioneermodels} reviews the design of the latest MLP pioneering models, describes the differences and connections between Token-mixing MLP, convolution, and self-attention mechanism, and presents the bottlenecks and challenges faced by the seemly new paradigm. Section~\ref{sec:blocks} and Section~\ref{sec:stages} discuss the block evolution and network architecture of MLP-like variants. Section~\ref{sec:performance} sheds light on applications of MLP-like variants. Section~\ref{sec:Outlook} gives our summary and discusses potential future research directions. 

\section{Preliminary}
\label{sec:preliminary}
For completeness and to provide helpful insight into the visual deep MLP presented in the subsequent sections, we briefly introduce MLP, CNN, and Transformer, including their brief histories and corresponding paradigms.

\subsection{Multilayer Perceptron and Boltzmann Machine}
The original “Perceptron” model was developed by Frank Rosenblatt in 1958~\cite{rosenblatt1958perceptron}, which can be viewed as a fully connected layer with only one output element. In 1969, a famous book entitled \emph{Perceptrons} by Marvin Minsky and Seymour Papert~\cite{minsky1969perceptrons} critically analyzed perceptron and pointed out several critical weaknesses of perceptron, including that perceptron was unable to learn an XOR function. For a while, interest in perceptron waned.

Interest in perceptron revived in 1985 when Hinton \emph{et al}.~\cite{rumelhart1985learning} recognized that a feed-forward neural network with two or more layers had a greater fitting ability than a single-layered perceptron. Indeed Hinton \emph{et al.} proposed the multilayer perceptron (MLP), a network composed of multiple layers of perceptrons and activation function, to solve the XOR problem. And they provided a reasonably effective training algorithm called \emph{backpropagation} for neural networks. As shown in 1989 by Cybenko, Hornik \emph{et al}., and Funahash~\cite{cybenko1989approximation,hornik1989multilayer,funahashi1989approximate}, MLPs are universal function approximators and can be used to construct mathematical models for classification and regression analysis.

In MLP, the fully connected layer can be viewed as a paradigm for extracting features. As the name implies, the main feature lies in the full connectivity, i.e., all the neurons from one layer are connected to every neuron in the following layer (Figure~\ref{fig:paradigmConpare-mlp}). One problem with fully connected layers is the input resolution sensitivity, where the number of neurons is related to the input size. Another significant problem with full connectivity is the enormous parameter cardinality and computational cost, growing quadratically with the image resolution.

The Boltzmann machine~\cite{ackley1985learning} proposed in 1985 is more theoretically intriguing, because of the analogy of its dynamics to simple physical processes. And a restricted Boltzmann machine (RBM)~\cite{smolensky1986information} comprises a layer of visible neurons and a layer of hidden neurons with only visible-hidden connections between the two layers. RBM is stacked~\cite{hinton2006reducing} and driven by recovering to a minimum energy state computed by a predetermined energy function. Contrarily speaking, stacked RBM is more computationally expensive than MLP, especially during inference. Computing power has been the main factor limiting the development of both MLP and RBM.

\subsection{Convolution Neural Network}
\label{subsec:conv_mechanism}
CNN was first proposed by Fukushima~\cite{fukushima1979neural,DBLP:journals/pr/FukushimaM82,DBLP:journals/tsmc/FukushimaMI83} in an architecture called Neocognitron, which involved multiple pooling and convolutional layers, inspired later CNNs. In 1989, LeCun \emph{et al}.~\cite{lecun1989backpropagation} proposed a multilayered CNN for handwritten zip code recognition, and the prototype of the architecture later called LeNet. After years of 
research, LeCun~\cite{lecun1998gradient} proposed LeNet-5, which outperformed all other models on handwritten character recognition. In the CPU era, it was widely accepted that the back-propagation algorithm was ineffective, considering the limited computational power of CPUs, in converging to the global minima of the error surface, and hand-crafted features were generally better than that of CNN-based extractors~\cite{schmidhuber2007new}.

In 2007, NVIDIA developed the CUDA programming platform~\cite{nickolls2008scalable,DBLP:journals/micro/LindholmNOM08}, and in 2009, ImageNet~\cite{deng2009imagenet}, a large image dataset, was proposed to provide the raw material for networks to learn image features autonomously. Three years later, AlexNet~\cite{krizhevsky2012imagenet} won the ImageNet competition, a symbolically significant event of the first paradigm shift. CNN-based architectures have gradually been utilized to extract image features automatically instead of hand-crafted features. In traditional computer vision algorithms, features such as gradient, texture, and color are extracted locally. Hence, the inductive biases inherent to CNNs, such as local connectivity and translation invariance, significantly help image feature extraction. The development of self-supervision~\cite{zhang2018local,chen2020simple,he2020momentum,chen2020improved} and training strategies~\cite{he2019bag,kolesnikov2020big,rw2019timm,mmseg2020,mmdetection,DBLP:journals/pami/YuTZRT22} further assisted the continuous improvement of CNNs. In addition to classification, CNNs outperform traditional algorithms for almost all computer vision tasks such as object detection~\cite{ren2015faster,redmon2018yolov3}, segmentation~\cite{he2017mask,ronneberger2015u}, demosaicing~\cite{syu2018learning}, super-resolution~\cite{dong2014learning,dong2016accelerating,yu2018wide}, and deblurring~\cite{nah2017deep}. \emph{CNN is the de-facto standard for computer vision} has become the consensus of the vision community.

\begin{figure}
    \begin{minipage}{0.5\textwidth}
        \subfigure[Convolution]{  			 
            \includegraphics[width=7.5cm]{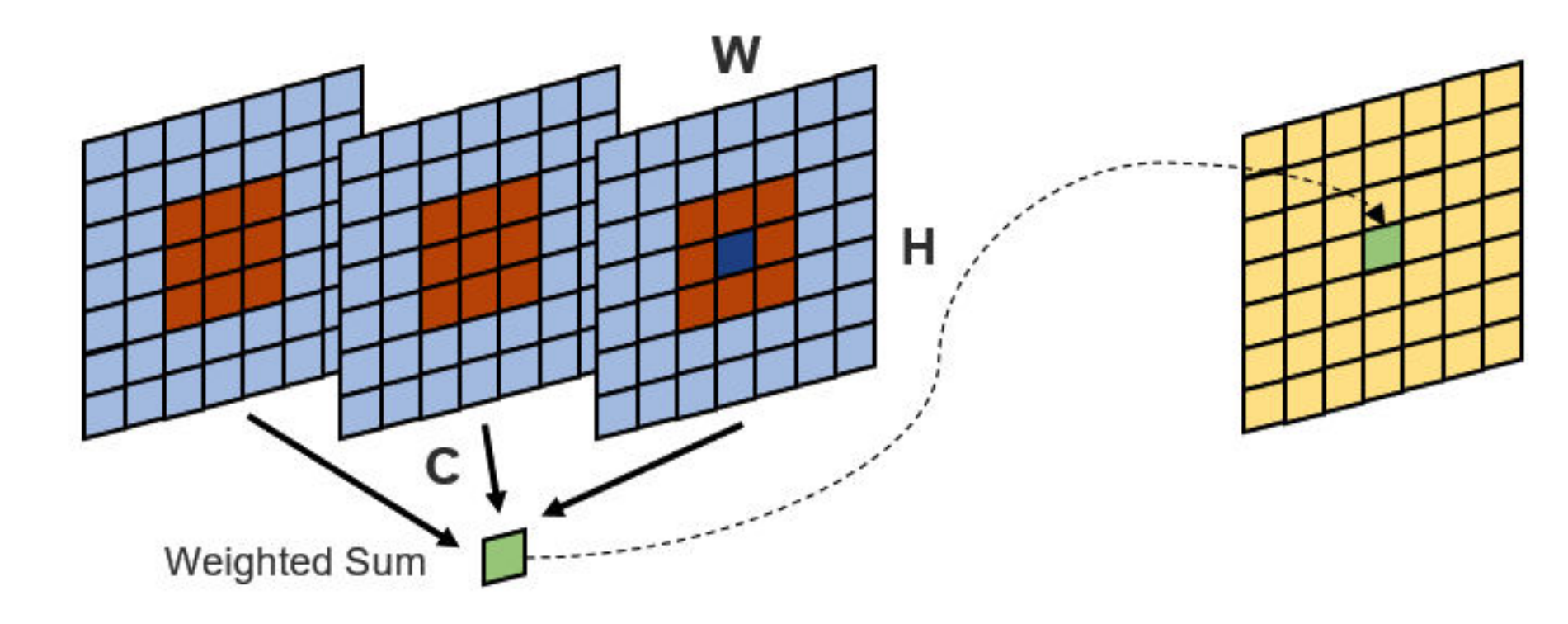}
            \label{fig:paradigmConpare-conv}
        }
        \subfigure[Depthwise Convolution]{   	 		 
            \includegraphics[width=7.5cm]{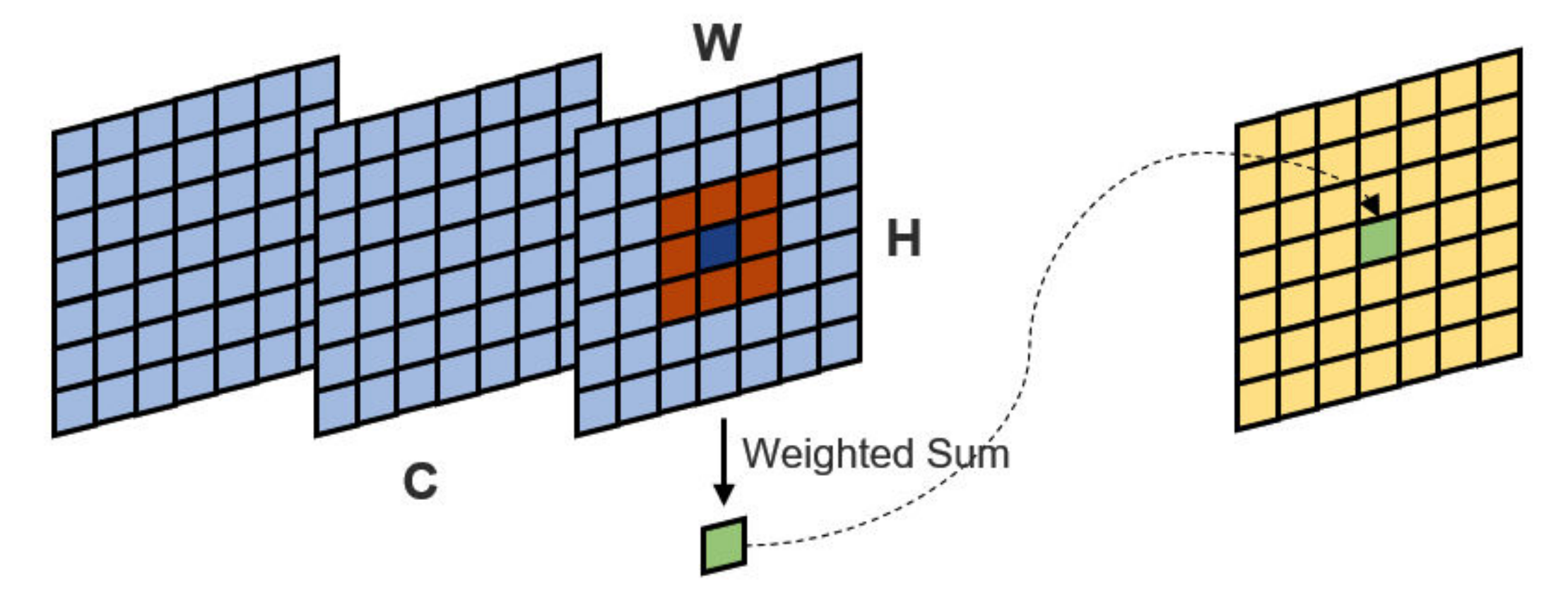}
            \label{fig:paradigmConpare-dwconv}
        } 
    \end{minipage}%
    \begin{minipage}{0.5\textwidth}
        \subfigure[Self-Attention]{   	 		 
            \includegraphics[width=7.5cm]{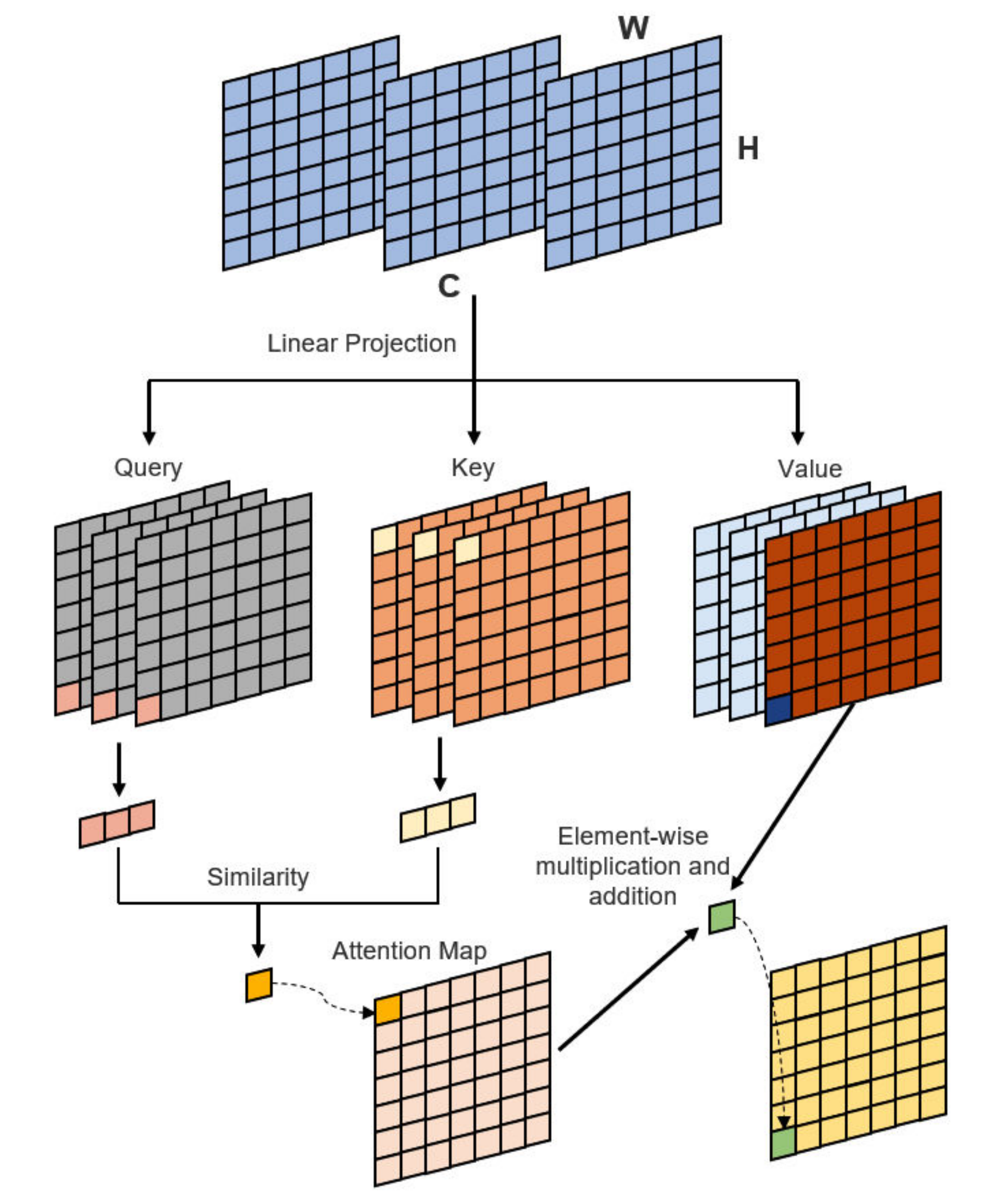}
            \label{fig:paradigmConpare-attention}
        } 
    \end{minipage}
    \subfigure[Conventional Fully Connected Layer]{   	 		 
        \includegraphics[width=7.5cm]{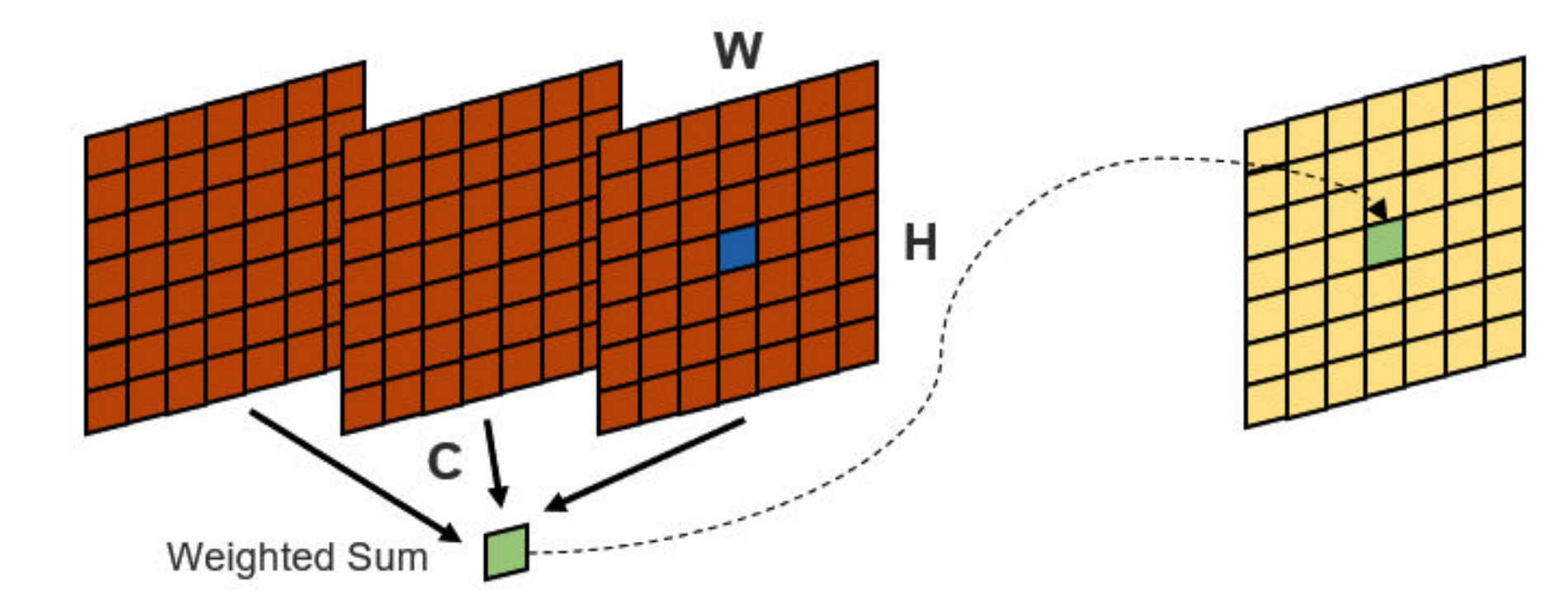}
        \label{fig:paradigmConpare-mlp}
    } 
    \subfigure[Token-mixing MLP]{   	 		 
        \includegraphics[width=7.5cm]{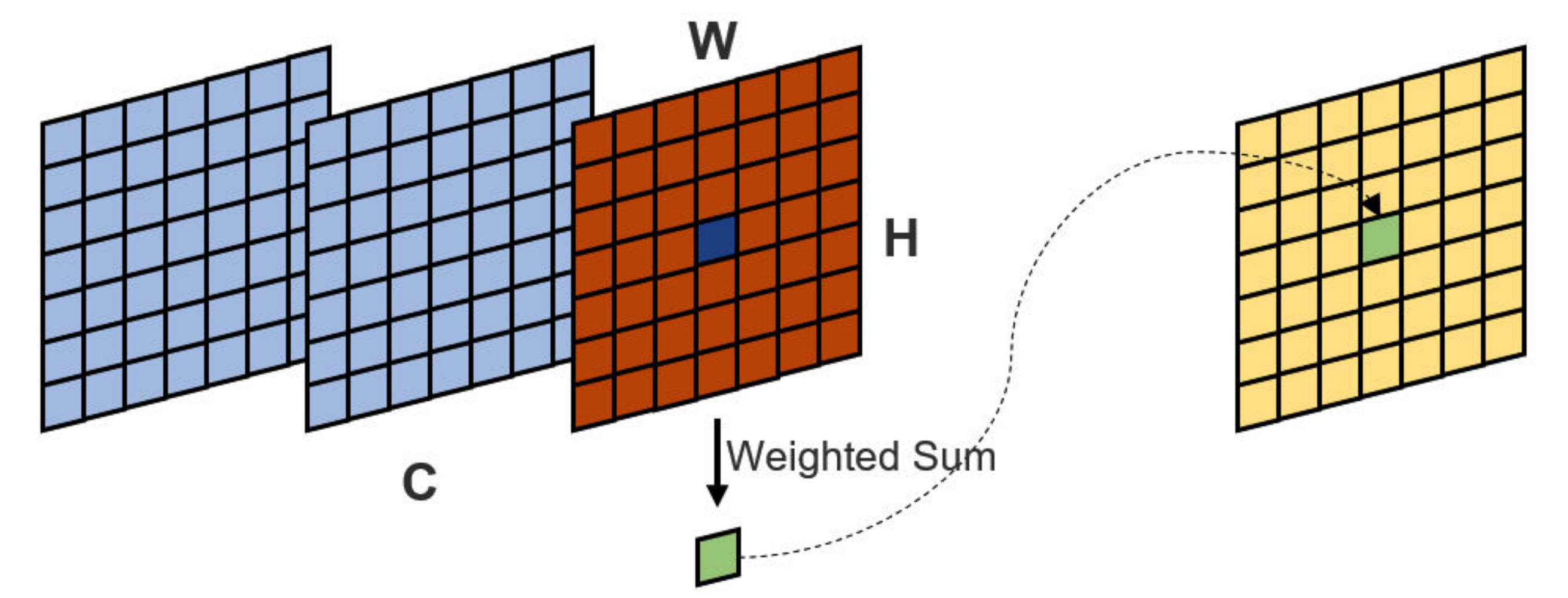}
        \label{fig:paradigmConpare-tkmlp}
    } 
    \caption{
        Illustrative shift between different weighted-sum paradigms. The input feature map is $H \times W \times C$, where $H$, $W$, $C$ are the feature map's height, width, and channel numbers, respectively. The light blue part highlights the input features, and the yellow is the output features. The dark blue dot represents the position of interest, the dark orange denotes other features used in the calculation process, and the green dot represents the corresponding output feature. The Token-mixing MLP is reduced to one fully connected layer to facilitate understanding. \emph{Linear Projection} performs an $1\times1$ convolution along the channel dimension, and \emph{Weighted Sum} means the elements are multiplied by the weights and then summed.
    }
    \label{fig:paradigmConpare}
\end{figure}

A CNN architecture typically comprises alternate convolutional and pooling layers with several fully connected layers behind, where the standard local-connected convolutional layer is the paradigm (Figure~\ref{fig:paradigmConpare-conv}). And depthwise convolution is a variant of convolution, where it applies different convolutional filters to different single channel (Figure~\ref{fig:paradigmConpare-dwconv}). In the convolutional operation, sliding kernels with the same set of weights can extract full sets of features within an image, making the convolutional layer more parameter-efficient than the fully connected layer.

\subsection{Vision Transformer}
\label{subsec:attention_mechanism}
Keeping pace with Moore’s law~\cite{moore1965cramming}, the computer capability has increased steadily with each new generation of chips. In 2020, visual researchers noted the application and success of the Transformer~\cite{vaswani2017attention} in natural language processing and suggested moving beyond the limits of local connectivity and towards global connectivity. Vision Transformer (ViT~\cite{dosovitskiy2020image}) is the first work promoting the research on Transformer in the field of vision. It uses stacked transformer blocks of the same architecture to extract visual features, where each transformer block comprises two parts: a multi-head self-attention layer and a two-layer MLP, in which layer normalization and residual path are also added. Since then, the Transformer-based architecture has been widely accepted in the vision community, outperforming CNN-based architectures in tasks such as objection detection~\cite{carion2020end,liu2021swin} and segmentation~\cite{liu2021swin,hatamizadeh2021unetr,valanarasu2021medical} and achieving state-of-the-art performance on denoising~\cite{chen2021pre}, deraining~\cite{chen2021pre}, super-resolution~\cite{chen2021pre,cao2021video,liang2021swinir,wang2021uformer}. Furthermore, several works~\cite{aldahdooh2021reveal,bhojanapalli2021understanding,mahmood2021robustness,naseer2021intriguing} demonstrate that the Transformer-based architecture is more robust than CNN-based methods. All the developments over the last two years indicate that \emph{Transformer has become another de-facto standard for computer vision}.

The paradigm in Transformer can be boiled down to the self-attention layer, where each input vector is linearly mapped into three different vectors: called \emph{query} $\mathbf{q}$, \emph{key} $\mathbf{k}$ and \emph{value} $\mathbf{v}$. Then the \emph{query}, \emph{key} and \emph{value} vectors come from different input vectors are aggregated into three corresponding matrices, denoted, $\mathbf{Q}$, $\mathbf{K}$ and $\mathbf{V}$ (Figure~\ref{fig:paradigmConpare-attention}). The self-attention mainly computes the cosine similarity between each query vector and all key vectors, and the $\operatorname{SoftMax}$ is applied to normalize the similarity and obtain the attention matrix. Output features then become the weighted sum of all value vectors in the entire space, where the attention matrix gives the weights. Compared with the convolutional layer, which focuses only on local characteristics, the self-attention layer can capture long-distance characteristics and easily derive global information.

\section{Pioneering Model and New Paradigm}
\label{sec:Pioneermodels}

The success of Vision Transformer marks the paradigm shift to the era of the global receptive field in computer vision, placing the consecutive question "Can we further abandon the artificially designed self-attention mechanism and let the model learn the global weights matrix autonomously from the raw data?". This motivation reminds researchers of the long-dusted simplest structure, MLP. After a long period of slumber, MLP finally reappears in May 2021, when the first deep MLP, called MLP-Mixer~\cite{tolstikhin2021mlp}, is launched.

This section reviews in detail the structure of the latest so-called pioneering MLP model, MLP-Mixer~\cite{tolstikhin2021mlp}, followed by a brief review of the contemporaneous ResMLP~\cite{touvron2021resmlp} as well as the FeedForward~\cite{melas2021you}. After that, we strip the new paradigm, Token-mixing MLP from the network, and elaborate its differences and connections with convolution and self-attention mechanisms. Finally, we explore the bottlenecks of Token-mixing MLP and lay the foundation for introducing subsequent variants.

\subsection{Structure of Pioneering Model}
\begin{figure}[!htb]
  \begin{minipage}[b]{1.0\linewidth}
    \centering
    \centerline{\includegraphics[width=\linewidth]{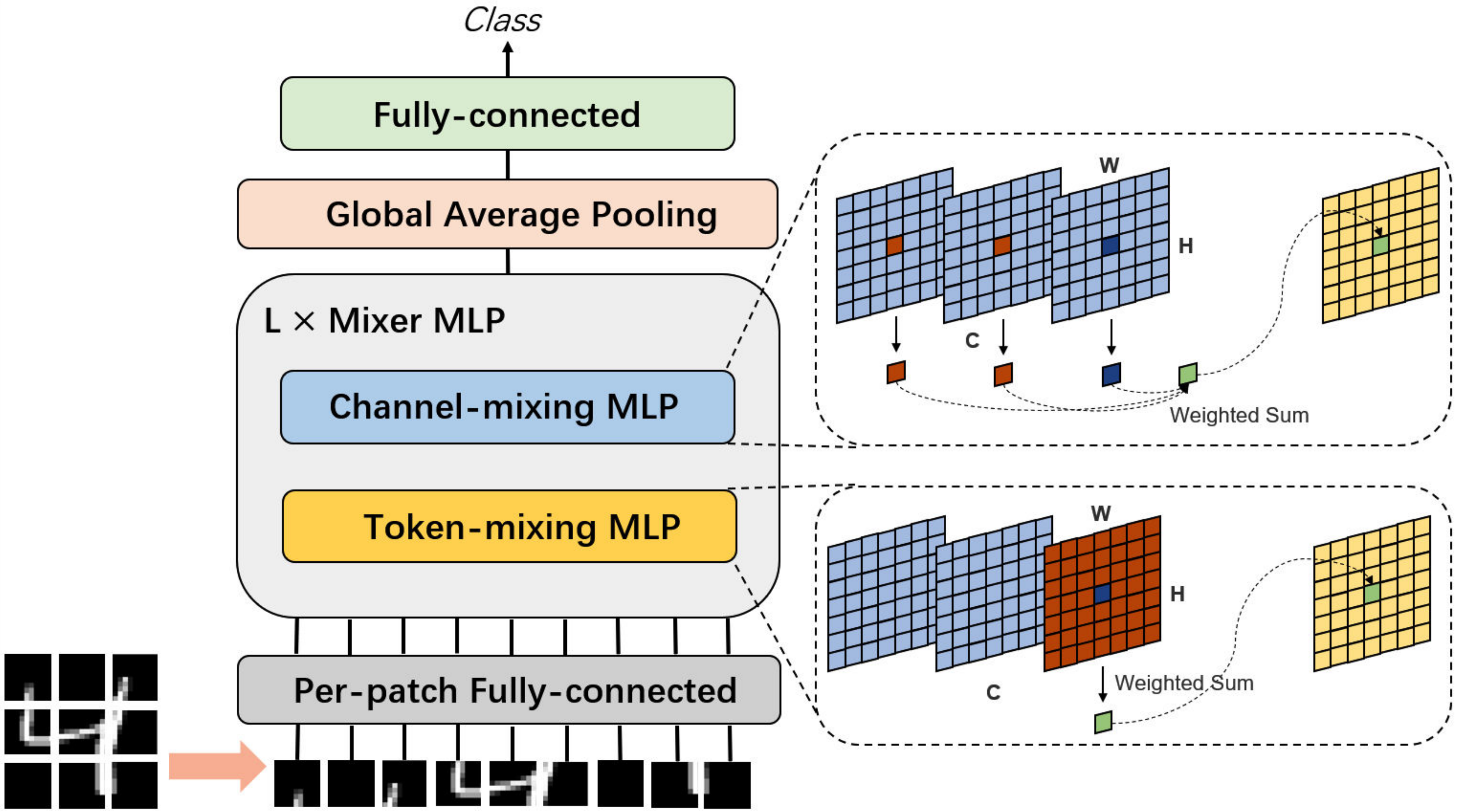}}
  \end{minipage}
  \caption{
    MLP-Mixer structure. We omit the Layer Normalization, non-linear activation function, and the residual path to improve readability. The Token-mixing and Channel-mixing MLP are reduced to one fully connected layer to ease understanding. The expression in the dashed box is consistent with Figure~\ref{fig:paradigmConpare}.
  }
  \label{fig:mlpmixer}
\end{figure}

MLP-Mixer~\cite{tolstikhin2021mlp} is the first proposed visual deep MLP network identified by the vision community as the pioneering MLP model. Compare to conventional MLP, it gets deeper and involves several differences. In detail, the MLP-Mixer comprises three modules, a per-patch fully-connected layer for patch embedding, a stack of $L $ Mixer layers for feature extraction, and a classification header for classification (Figure~\ref{fig:mlpmixer}).

\textbf{Patch Embedding}. The patch embedding is inherited from ViT~\cite{dosovitskiy2020image}, comprising three steps: (1) Cut an image into non-overlap patches, (2) Flatten the patches, and (3) Linearly project these flattened patches. Specifically, an input image of the size $H \times W \times 3$ is splited into $S$ non-overlapping $p \times p \times 3$ patches, where $H$ and $W$ are the image's height and width, respectively, $S$ donates patch number, and $p$ represents the patch size (typically $14$ or $16$). The patch is also called a \emph{token}. Each patch is then unfolded into a vector $\mathbf{p} \in \mathbb{R}^{3p^2}$. In total, we obtain a set of flattened patches $\mathbf{P}=\left\{\mathbf{p}_{1}, \cdots, \mathbf{p}_{S}\right\}$, which are input of the MLP-Mixer. For each $\mathbf{p}_{i} \in \mathbf{P}$, the per-patch fully-connected layer maps it into a $C$-dimensional embedding vector:
\begin{equation}
  \begin{aligned}
    \begin{array}{ll}
        \mathbf{x}_i=\mathbf{W}\mathbf{p}_i^T, & \mathbf{W} \in \mathbb{R}^{C\times3p^2}, \mathbf{p}_i \in \mathbb{R}^{3p^2}, \text { for } i=1 \ldots S,
    \end{array}
  \end{aligned}
\end{equation}
where $\mathbf{x}_i \in \mathbb{R}^{C}$ is the embedding vector of $\mathbf{p}_i$ and $\mathbf{W}$ represents weights of the per-patch fully-connected layer. In practice, it is possible to combine three steps presented above into a single step using a 2D convolution operation, where the convolutional kernel size and stride are equal to patch size.

\textbf{Mixer Layers}. The MLP-Mixer stacks $L$ Mixer Layers of the same architecture, where a single Mixer Layer essentially consists of a Token-mixing MLP and a Channel-mixing MLP. Let the patch features at the input of each Mixer Layer be $\mathbf{X}=\left[\mathbf{x}_{1}, \cdots, \mathbf{x}_{S}\right] \in \mathbb{R}^{C \times S}$, where $S$ is the number of patches, and $C$ is the dimension of each patch feature, i.e., the number of channels. The Token-mixing MLP works on each column of $\mathbf{X}$, maps $\mathbb{R}^{S} \mapsto \mathbb{R}^{S}$, and the weights are shared among all columns. The Channel-mixing MLP works on each row of $\mathbf{X}$, maps $\mathbb{R}^{C} \mapsto \mathbb{R}^{C}$, and the weights are shared among all rows. Both Token-mixing and Channel-mixing MLPs comprise two fully connected layers, and there is a non-linear activation function between the two layers. Thus, a Mixer Layer can be written as follows (omitting layer indices):
\begin{equation}
  \begin{aligned}
    \begin{array}{ll}
        \mathbf{U}_{*, i}=\mathbf{X}_{*, i}+\mathbf{W}_{2} \sigma\left(\mathbf{W}_{1} \operatorname{LayerNorm}(\mathbf{X})_{*, i}\right), & \text { for } i=1 \ldots C, \\
        \mathbf{Y}_{j, *}=\mathbf{U}_{j, *}+\mathbf{W}_{4} \sigma\left(\mathbf{W}_{3} \operatorname{LayerNorm}(\mathbf{U})_{j, *}\right), & \text { for } j=1 \ldots S ,
    \end{array}
  \end{aligned}
\end{equation}
where $\sigma$ is the GELU~\cite{hendrycks2016gaussian} activation function, and $\operatorname{LayerNorm}(\cdot)$ denotes the layer normalization~\cite{ba2016layer} widely used in Transformer-based models. $\mathbf{W}$ represents the weights of a fully connected layer, where $\mathbf{W}_{1} \in \mathbb{R}^{rS \times S}$, $\mathbf{W}_{2} \in \mathbb{R}^{S \times rS}$, $\mathbf{W}_{3} \in \mathbb{R}^{rC \times C}$ and $\mathbf{W}_{4} \in \mathbb{R}^{C \times rC}$, and $r > 1$ is the expansion ratio (commonly $r=4$). It is worth mentioning that each Mixer Layer takes an input of the same size, which is most similar to Transformers or deep Recurrent Neural Networks (RNNs) in other domains. However, it opposes most CNNs, which have a \emph{pyramidal} structure: deeper layers have a lower resolution input but more channels.

\textbf{Classification Header}. After processing with $L$ stacking Mixer layers, $S$ patch features are generated. Then, a global vector is calculated from the features through the average pooling scheme, which is forwarded into a fully connected layer for classification.

Compared to MLP-Mixer, FeedForward~\cite{melas2021you} and ResMLP~\cite{touvron2021resmlp} are proposed a few days later. FeedForward~\cite{melas2021you} adopts essentially the same structure as the MLP-Mixer, but swaps the Channel-mixing MLP and Token-mixing MLP order. As another contemporary work, ResMLP~\cite{touvron2021resmlp} simplifies the Token-mixing MLP in the MLP-Mixer from two fully connected layers to one. Meanwhile, ResMLP proposes an affine element-wise transformation to replace the layer normalization in the MLP-Mixer and stabilize training.

Experimentally, these MLP pioneering models achieve comparable performance to CNN and ViT for image classification (Table~\ref{table:ImageNet}). These empirical results significantly break past perceptions, challenge the necessity for the convolutional layer and the self-attention layer, and prompt the community to rethink whether the convolutional layer or the self-attention layer is necessary. The latter induces us to explore whether a pure MLP stack will become the new paradigm for computer vision.

\subsection{Token-mixing MLP, a New Paradigm?}

In order to find out whether pure MLP stacked into a Mixer Layer will become a new paradigm, it is necessary first to reveal the difference between it, convolution, and self-attention mechanisms. There is an indisputable fact that both the Channel-mixing MLP in MLP-Mixer and the MLP in ViT are just a $1 \times 1$ convolution commonly used in CNNs, allowing communication between different channels. The core points to compare come naturally to Token-mixing MLP, self-attention, and convolution, which allow communication between different spatial locations. A detailed comparison between the three is reported in Table~\ref{table:compare1}.

Convolution usually performs the aggregation computation of spatial information in a local region, but poorly models long-term dependencies. The Token-mixing MLP (Figure~\ref{fig:paradigmConpare-tkmlp}) can be viewed as an unusual convolution type whose convolutional kernel covers the entire space. To enhance efficiency, it aggregates spatial information from a single channel and shares weights for all channels. This is very close to the depthwise convolution~\cite{chollet2017xception} (Figure~\ref{fig:paradigmConpare-dwconv}) used in CNN, which independently applies convolutions to each channel. However, the convolutional kernels in depthwise convolution are usually small and not shared between each channel. The self-attention mechanism considers all patches as well, where the weights are dynamically generated based on the input, while weights in Token-mixing MLP and the convolutional layer are fixed and input-independent.

\begin{table}[]
\centering
\caption{Comparison between convolution, self-attention, and Token-mixing MLP. $H$, $W$, $C$ are the height, width and channel numbers of the feature map, respectively. $k$ is the convolutional kernel size. \emph{Information Aggregation} refers to whether the weights are fixed or dynamically generated based on the input during inference. \emph{Resolution Sensitive} refers to whether the operation is sensitive to input resolution. \emph{Spatial} refers whether feature extraction is sensitive to the spatial location of objects, \emph{specific} means true, while \emph{agnostic} means false. \emph{Channel-specific} means no weights are shared between channels, \emph{Channel-agnostic} means weights are shared between channels.}
\resizebox{\textwidth}{!}{
\begin{tabular}{l|ccccccc}
    \toprule
    Operation & \begin{tabular}[c]{@{}c@{}}Information\\ Aggregation\end{tabular} & \begin{tabular}[c]{@{}c@{}}Receptive\\ Field\end{tabular} & \begin{tabular}[c]{@{}c@{}}Resolution\\ Sensitive\end{tabular} & Spatial & Channel & Params & FLOPs \\ 
    \midrule
    Convolution & static & local & False & agnostic & specific & $\mathcal{O}(k^2C^2)$ & $\mathcal{O}(HWC^2)$ \\
    Depthwise Convolution & static & local & False & agnostic & specific & $\mathcal{O}(k^2C)$ & $\mathcal{O}(HWC)$ \\
    Self-Attention~\cite{dosovitskiy2020image} & dynamic & global & False & agnostic & specific & $\mathcal{O}(3C^2)$ & $\mathcal{O}(H^2W^2C)$ \\
    Token-mixing MLP~\cite{tolstikhin2021mlp} & static & global & True & specific & agnostic & $\mathcal{O}(H^2W^2)$ & $\mathcal{O}(H^2W^2C)$ \\
    \bottomrule
\end{tabular}
}
\label{table:compare1}
\end{table}

We now shift to another important metric, namely complexity. Without loss of generality,  we assume that the input feature map size is $HW \times C$ (or $H \times W \times C$ in CNNs), where $H$ and $W$ are the spatial resolutions, and $C$ is the number of channels. Intuitively, local computation has minimal computational complexity, i.e., $\mathcal{O}(HWC)$ in depthwise convolution and $\mathcal{O}(HWC^2)$ in dense convolution. However, both self-attention and Token-mixing MLP involving a global receptive field have greater complexity, $\mathcal{O}(H^2W^2C)$. Fortunately, Token-mixing MLP is less computationally intensive than the self-attention layer due to the lack of calculations such as the attention matrix. As for the parameter cardinality, the parameter complexity is $\mathcal{O}(H^2W^2)$ in Token-mixing MLP, strongly correlated with the image resolution. So once the network is fixed, the corresponding image resolution is also fixed. In comparison, other paradigms have more advantages in parameters. The newly proposed MLP-like variants are optimized in complexity, see Sec.~\ref{subsec:complexity}.

\begin{figure}[!htb]    	
    \centering    	
    \subfigure[MLP-Mixer]{  			 
        \includegraphics[width=\linewidth]{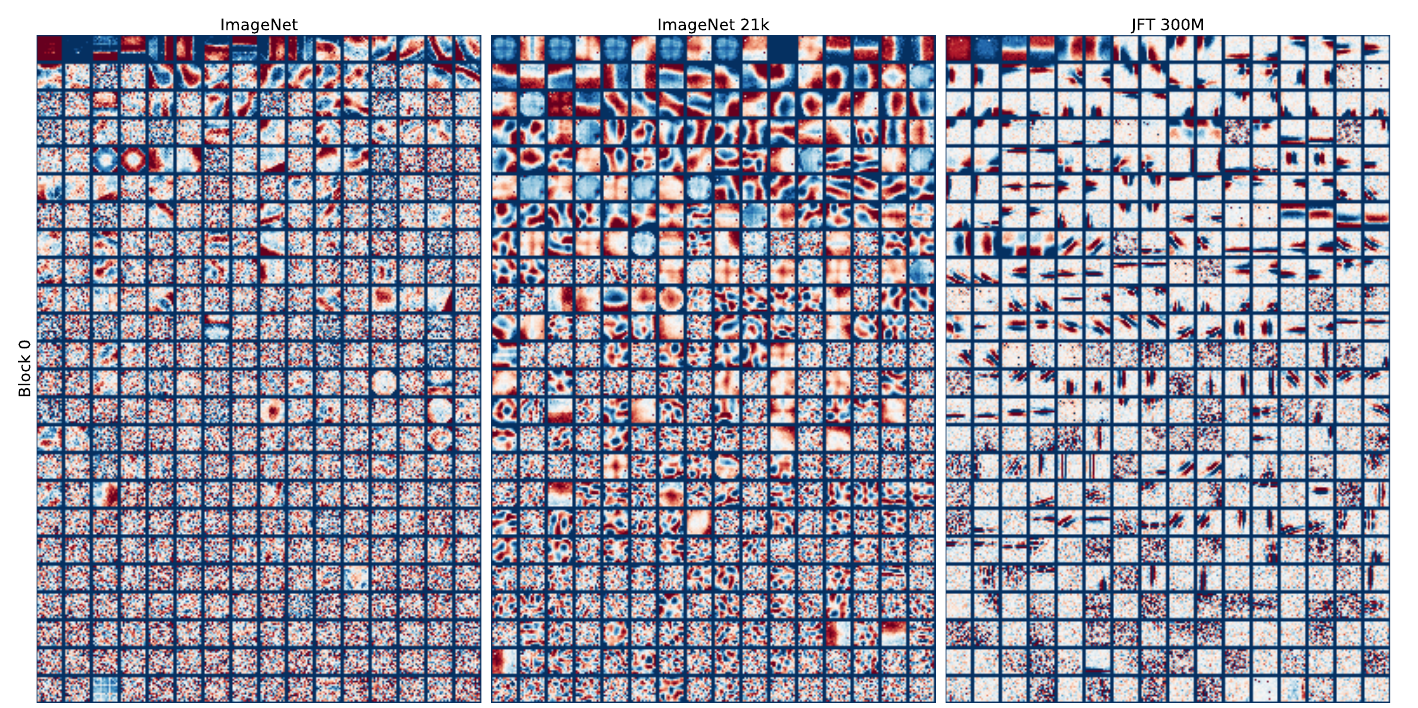}
    }    	 
    \subfigure[ResMLP]{   	 		 
        \includegraphics[width=\linewidth]{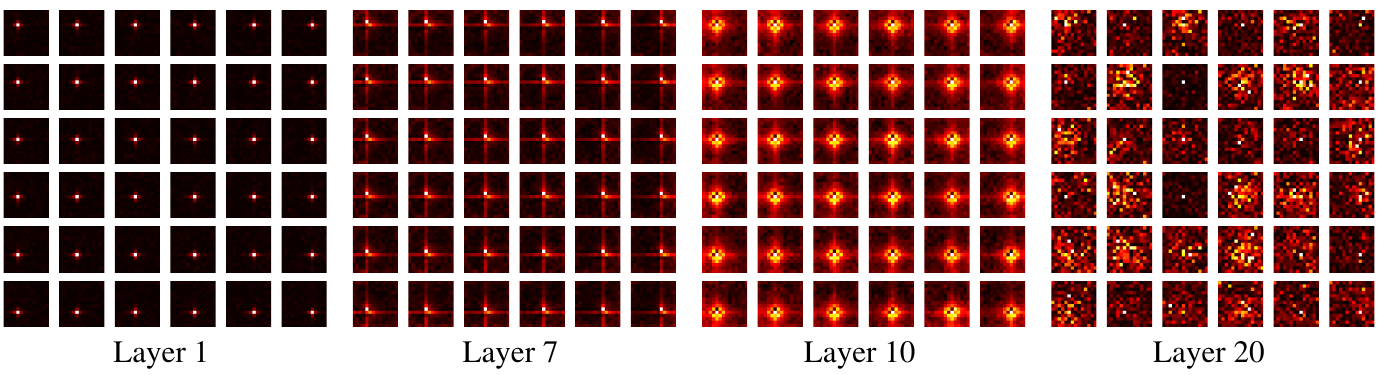}
    }
    \caption{Visualizing the weights of the fully connected layers in MLP-Mixer and ResMLP. Each weight matrix is resized to $14 \times 14$ pixel images. In MLP-Mixer, white denotes that weight is 0, red means positive weights, blue means negative weights, and the brighter, the greater the weight. In ResMLP, black indicates that weight is 0, and the brighter, the greater the weight's absolute value. The results are from~\cite{tolstikhin2021mlp} and~\cite{touvron2021resmlp}, respectively.}  
    \label{fig:filters} 
\end{figure}

The above comparative analysis reveals that Token-mixing MLP is seemly a new paradigm. But what does the new paradigm bring to the learning weights? Figure~\ref{fig:filters} visualizes the weights of fully connected layers (FC kernels) of MLP-Mixer and ResMLP trained on ImageNet~\cite{deng2009imagenet} or JFT-300M~\cite{sun2017revisiting}. ImageNet-1k contains 1.2 million labeled images, ImageNet-21k involves 14 million images, JFT-300M has 300 million images. As the amount of training data increases, the number of FC kernels for locality computation in MLP-Mixer increases. ResMLP's shallow FC kernels also present some local connectivity properties. These fully connected layers actually still perform local computation,which convolutional layers can replace. Thus, we conclude that \textbf{The shallow fully connected layers of deep MLP implement convolution-like scheme}. As the number of layers increases, i.e., the network becomes deeper, the effective range of receptive field increases, and the weights become disorganized. However, it is unclear if this is due to a lack of training data or if it should be so. Notably, while MLP-Mixer and ResMLP share a highly similar structure (except for the normalization layer), their learning weights are vastly different. This questions whether MLP is learning generic visual features. Moreover, MLP's interpretability stands far behind.

\subsection{Bottlenecks}
\label{subsec:mlp-question}
Based on the above analysis and comparison, it is evident that the seemingly new paradigm still faces several bottlenecks:
\begin{enumerate}[(1)]
\item Without the inductive biases of the local connectivity and the self-attention, the Token-mixing MLP is more flexible and has a stronger fitting capability. However, it has a greater risk of over-fitting than the convolutional and self-attention layers. Thus, the large-scale dataset is needed to shorten the classification accuracy gap between MLP-Mixer, ViT, and CNN.
\item The complexity of the Token-mixing MLP is quadratic to image size, which, for the current computing capability, makes it intractable for existing MLP-like models on high-resolution images.
\item The Token-mixing MLP is resolution sensitive, and the network cannot deal with a flexible input resolution once the number of neurons in the fully connected layer is set. However, some tasks adopt a multi-scale training strategy~\cite{carion2020end} and have different input resolutions between training and evaluation stages~\cite{lin2014microsoft,cordts2016cityscapes}. In these cases, MLP models are non-transferable and impractical.
\end{enumerate}

After several explorations and practices to address these challenges, the vision community has developed many MLP-like variants. Their main contributions are modifications to the Token-mixing MLP, including reducing computational effort, removing resolution sensitivity, and reintroducing local receptive fields. These variants will be described in detail in the subsequent sections.

\begin{figure}[!htb]
  \begin{minipage}[b]{1.0\linewidth}
    \centering
    \centerline{\includegraphics[width=\linewidth]{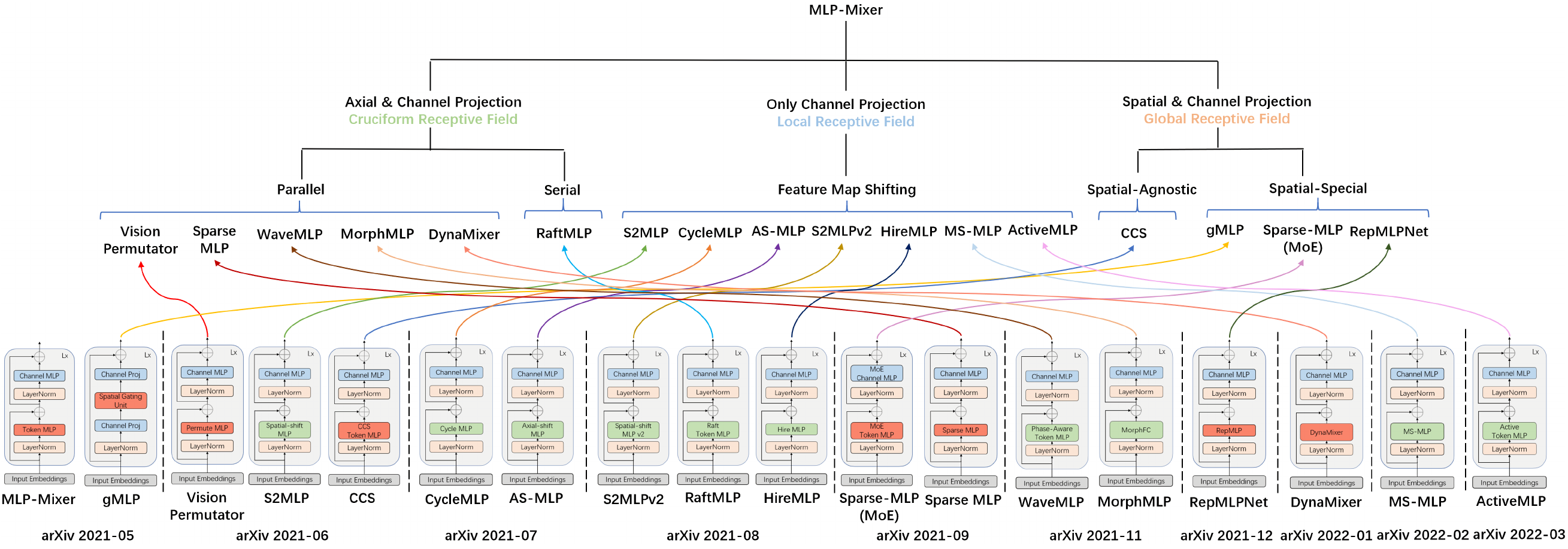}}
  \end{minipage}
  \caption{
    Block comparison of MLP-Mixer variants. Sorted from left to right by model release month (“arXiv 2021-05” denotes “May, 2021”). \textcolor{RedOrange}{\emph{RedOrange}} Token-mixing modules are resolution sensitive, while \textcolor{green}{\emph{Green}} Token-mixing modules are resolution insensitive.
  }
  \label{fig:networks}
\end{figure}

\section{Block of MLP Variants}
\label{sec:blocks}

In order to overcome the challenges faced by the MLP-Mixer, the vision community has made several attempts and proposed many MLP-like variants. The improvements focus on redesigning the network's interior parts, i.e., the block module. The lower part of Figure~\ref{fig:networks} illustrates the block designs of the latest MLP-like variants, highlighting that they are primarily modified for Token-mixing MLP. Except for the gMLP~\cite{liu2021pay}, the remaining blocks retain the tandem spatial MLP and the channel MLP. Moreover, most of the improvements reduce the spatial MLP's sensitivity to image resolution (Green Rectangle).

In this work, we reproduce most variants of MLP-like models in Jittor~\cite{hu2020jittor,jittormlp} and Pytorch~\cite{paszke2019pytorch}. Moreover, this section first details the redesigned blocks of the latest MLP-like variants, then compare their properties and receptive field, and finally discusses the findings.

\subsection{MLP Block Variants}

We divide the MLP block variants into three categories: (1) mappings employing both the axial direction and channel dimensions, (2) mappings considering only channel dimensions, and (3) mappings utilizing the entire spatial and channel dimensions. The upper part of Figure~\ref{fig:networks} categorizes the network variants. Since the Channel-mixing MLP is basically the same (except for the gMLP), we review and describe the changes to the Token-mixing MLP.

\subsubsection{Axial and channel projection blocks}

The global receptive field of the initial Token-mixing MLP is heavily parameterized and computationally very complex. Some researchers have proposed orthogonally decomposing the spatial projections and maintaining long-range dependence while no longer encoding spatial information along a flat spatial dimension.

\begin{figure}[!htb]    	
    \centering    	
    \subfigure[ViP]{  			 
        \includegraphics[width=7cm]{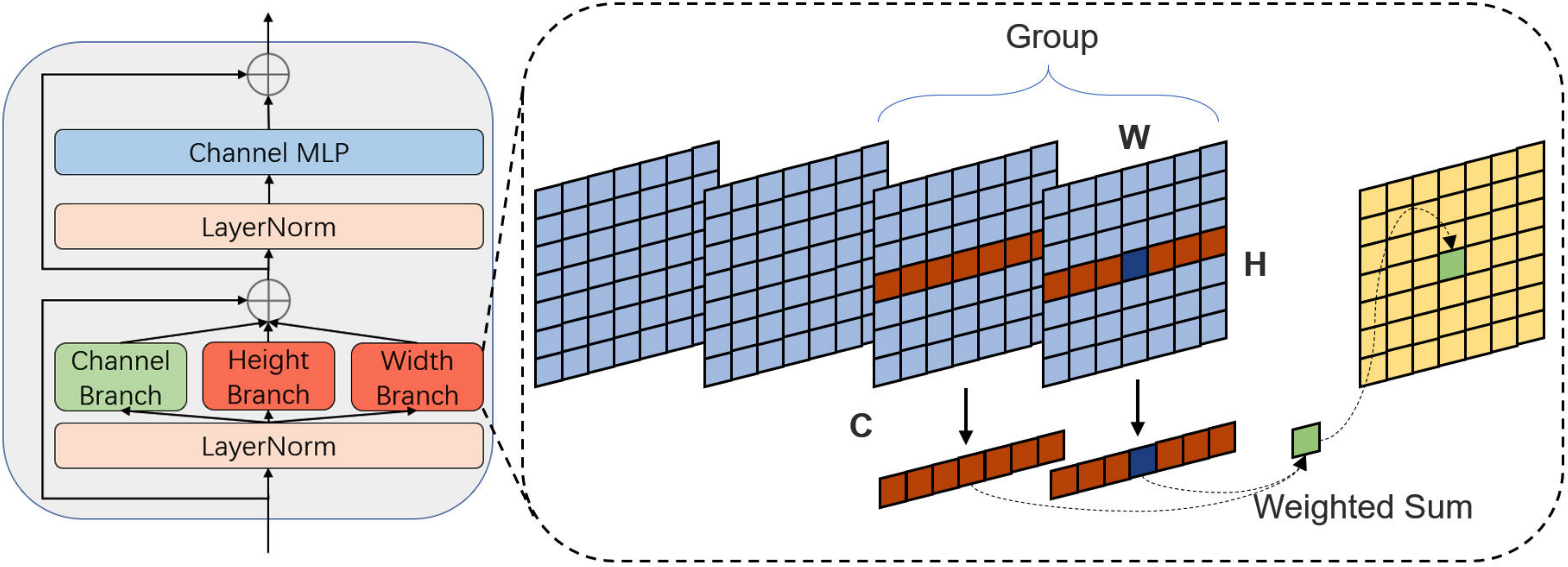}
        \label{fig:axial_channel_module_vip}
    }    	 
    \subfigure[Sparse MLP]{   	 		 
        \includegraphics[width=7cm]{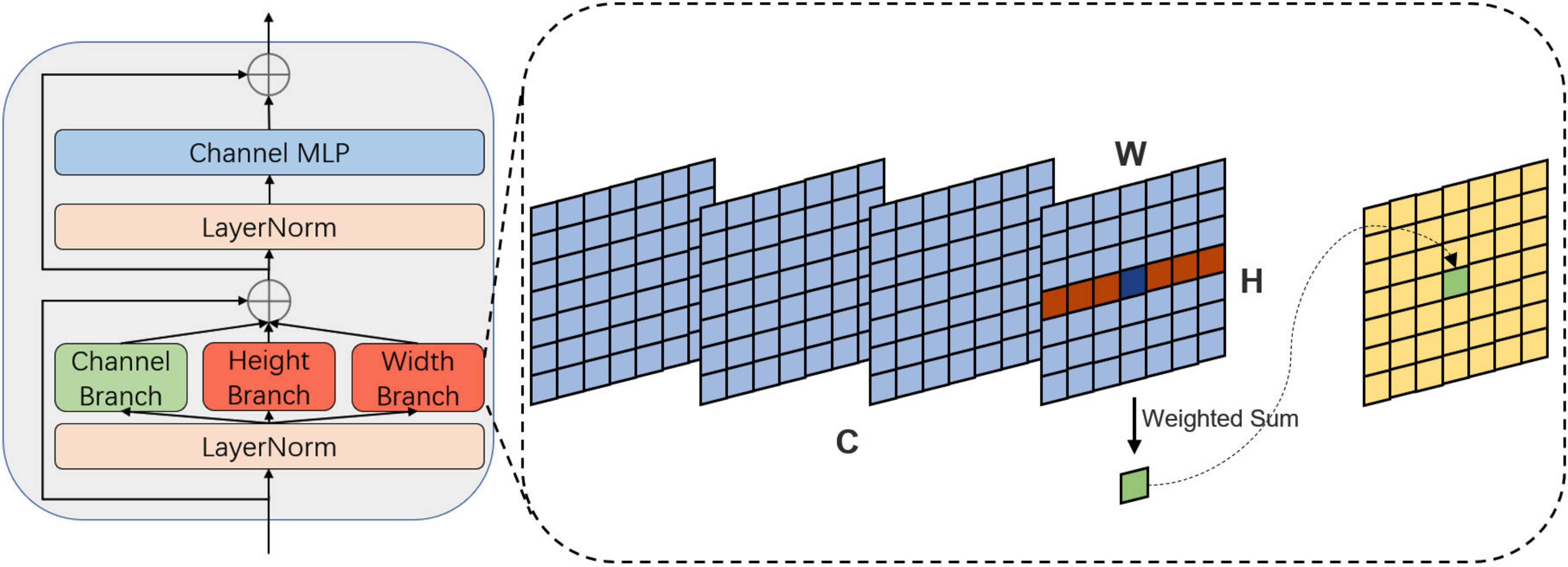}
        \label{fig:axial_channel_module_sparsemlp}
    }
    \subfigure[RaftMLP]{  			 
        \includegraphics[width=7cm]{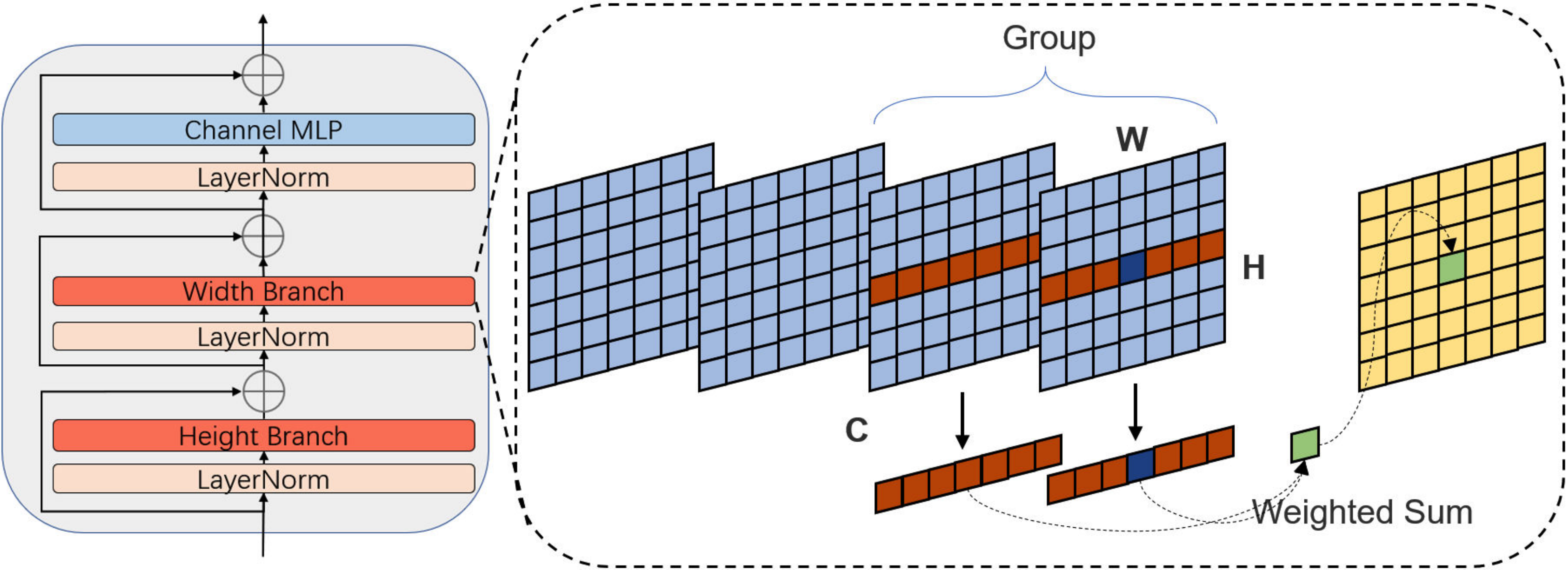}
        \label{fig:axial_channel_module_raftmlp}
    }    	 
    \subfigure[DynaMixer]{   	 		 
        \includegraphics[width=7cm]{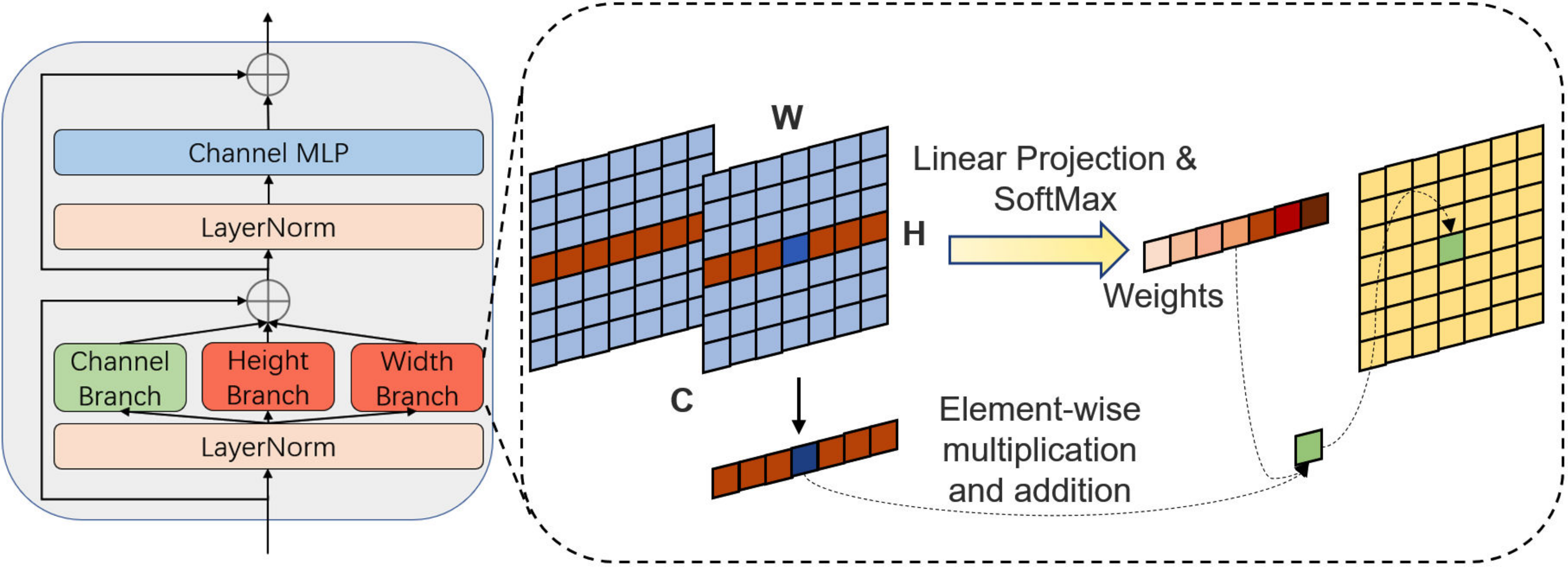}
        \label{fig:axial_channel_module_dynamixer}
    }
    \subfigure[WaveMLP]{  			 
        \includegraphics[width=7cm]{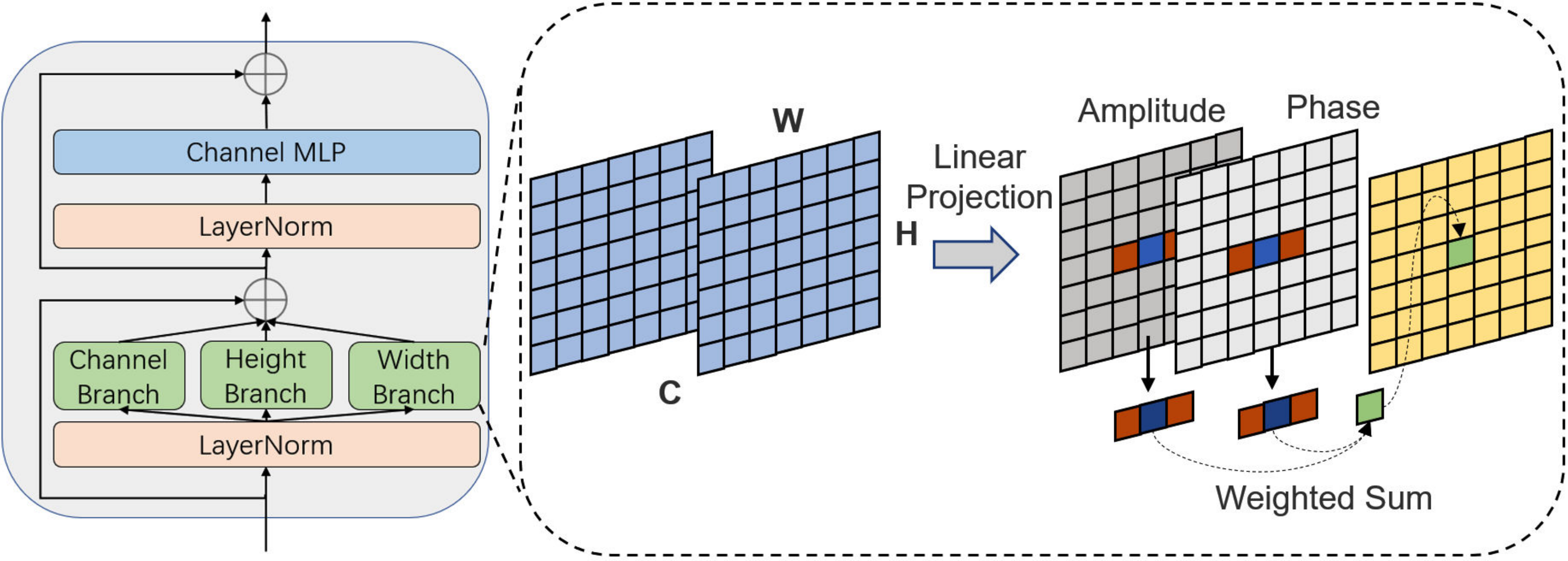}
        \label{fig:axial_channel_module_wavemlp}
    }    	 
    \subfigure[MorphMLP]{   	 		 
        \includegraphics[width=7cm]{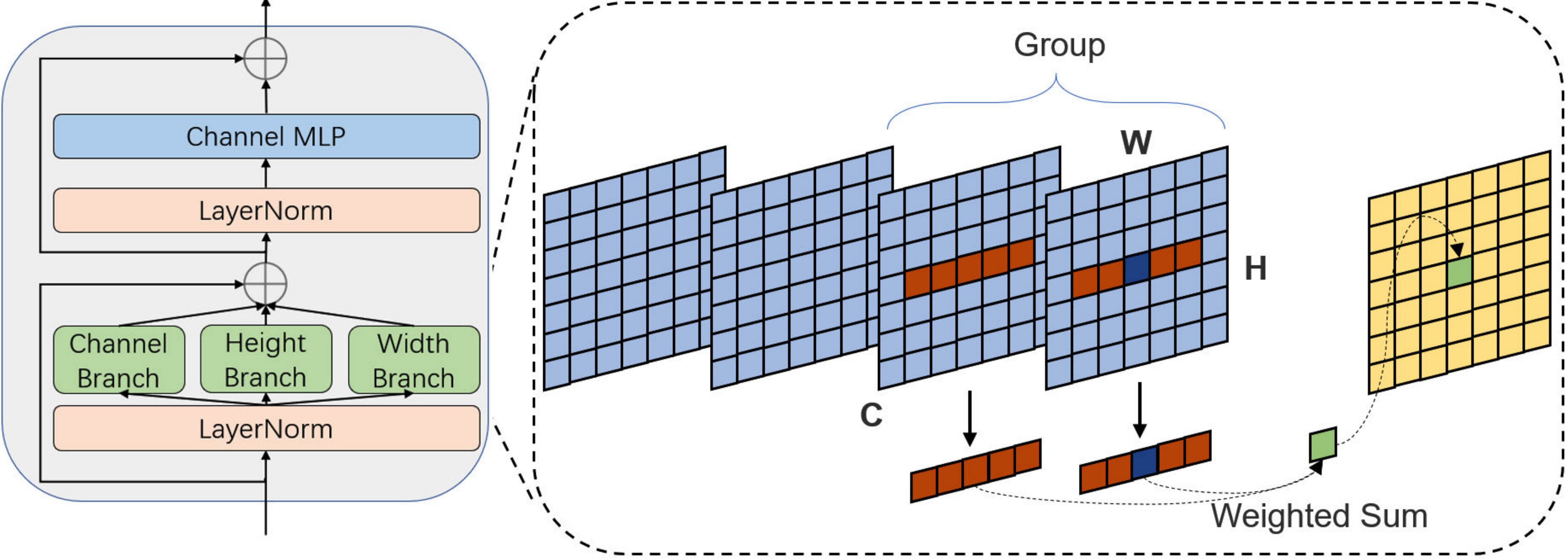}
        \label{fig:axial_channel_module_morphmlp}
    }
    \caption{Visualizing the module diagram of the MLP-like variants. Except for RaftMLP, the remaining block designs adopt three-branch parallelism. \textcolor{RedOrange}{\emph{RedOrange}} branch is image resolution sensitive, while \textcolor{green}{\emph{Green}} branch is resolution insensitive. The \emph{Channel Branch} and \emph{Linear Projection} both perform a linear projection along the channel dimension. "Group" means that the channel is split into different groups. The expression in the dashed box is consistent with Figure~\ref{fig:paradigmConpare}.}  
    \label{fig:axial_channel_module} 
\end{figure}

Hou \emph{et al.}~\cite{hou2021vision} present the Vision Permutator (ViP), which separately encodes the feature representations along the height and width dimensions with linear projections. This design allows ViP to capture long-range dependencies along one spatial direction while preserving precise positional information along the other direction. As illustrated in Figure~\ref{fig:axial_channel_module_vip}, Permute-MLP comprises three branches responsible for encoding information along the height, width, or channel dimension. Specifically, it first splits the feature map into $g$ segments along the channel dimension, where $g = C/H$ and the segments are then concatenated along the height dimension. Thus, it maps $\mathbb{R}^{gH} \mapsto \mathbb{R}^{gH}$ and is shared across the width and part channels. After mapping, the feature map is recovered to the original shape. The branch treatment in the width direction is the same as in the height direction. The third path is a simple mapping process along the channel dimension, which can also be regarded as a $1 \times 1$ convolution. Finally, the outputs from all three branches are fused by exploiting the split attention~\cite{zhang2020resnest}.

Tang \emph{et al.}~\cite{tang2021sparse} adopt a strategy consistent with ViP for spatial information extraction and build an attention-free network called Sparse MLP. As illustrated in Figure~\ref{fig:axial_channel_module_sparsemlp}, the block contains three parallel branches. The difference from ViP is no longer splitting the feature map along the channel dimension, but directly mapping $\mathbb{R}^{H} \mapsto \mathbb{R}^{H}$ and $\mathbb{R}^{W} \mapsto \mathbb{R}^{W}$ along the height and width dimension. Without the split attention, Sparse MLP's fusion strategy involves concatenating the three tributary outputs by channel and then passing them through a fully connected layer for dimensionality reduction. There is also a minor modification where Sparse MLP places a depthwise convolution in front of each block.

RaftMLP~\cite{tatsunami2021raftmlp} employs serial mappings in high and wide dimensions to form a Raft-Token-mixing block (Figure~\ref{fig:axial_channel_module_raftmlp}), which is different from the parallel branches of ViP and Sparse MLP. In the specific implementation, the Raft-Token-mixing block also splits the feature map along the channel dimension, consistent with ViP.

DynaMixer~\cite{wang2022dynamixer} generates mixing matrices dynamically for each set of tokens to be mixed by considering their contents. DynaMixer adopts the parallel strategy for a computational speedup and mixes tokens in a row-wise and column-wise way (Figure~\ref{fig:axial_channel_module_dynamixer}). The proposed DynaMixer operation performs dimensionality reduction and flattening first and then utilizes a linear function to estimate a $H \times H$ or $W \times W$ mixing matrix. $\operatorname{SoftMax}$ is performed on each row of the mixing matrix to obtain the mixing weights. The output equals the product of the mix weights and the input.

WaveMLP~\cite{tang2021image} considers the dynamic aggregation tokens and image resolution issues. It considers each token a wave with both amplitude and phase information (Figure~\ref{fig:axial_channel_module_wavemlp}). The tokens are aggregated according to their varying contents from different input images with the dynamically produced phase. There are two parallel paths to aggregate spatial information along the horizontal and vertical directions, respectively. To address the issue of sensitivity to image resolution, WaveMLP adopts a simple strategy that restricts the fully connected layers only to connect tokens within a local window~\cite{liu2021swin}. However, the local window limits long-range dependencies.

MorphMLP~\cite{zhang2021morphmlp} considers long-range and short-range dependencies while continuing the static aggregation strategy (Figure~\ref{fig:axial_channel_module_morphmlp}). It focuses on local details in the early stages of the network and gradually changes to long-term modeling in the later stages. The local window is used to solve the image resolution sensitivity problem, and the window size increases as the number of layers increases. The authors find that such a feature extraction model is beneficial for images and videos.

\textbf{Discussion}: ViP, Sparse MLP, RaftMLP, and DynaMixer encode spatial information along the axial direction instead of the entire plane, preserving the long-range dependence to a certain extent and reducing the complexity of parameters and the computational cost. However, they are still image resolution sensitive. WaveMLP and MorphMLP adopt a local window strategy but discard long-range dependencies. Furthermore, all those variants cannot mix tokens both globally and locally.

\subsubsection{Channel-only projection blocks}

The mainstream method adopts Swin's proposal~\cite{liu2021swin} and uses a Local Window to achieve resolution insensitivity. Another approach replaces all the spatial fully connected layers with channel projection, i.e., $1 \times 1$ convolution. However, it makes the tokens no longer interact with each other, and the concept of the receptive field disappears. To reintroduce the receptive field, many works align features at different spatial locations to the same channel by shifting (or moving) the feature maps and then interacting with spatial information through channel projection. Such an operation enables an MLP-like architecture to achieve the same local receptive field as a CNN-like architecture.

\begin{figure}[!htb]    	
    \centering    	
    \subfigure[S2MLP]{  			 
        \includegraphics[width=7cm]{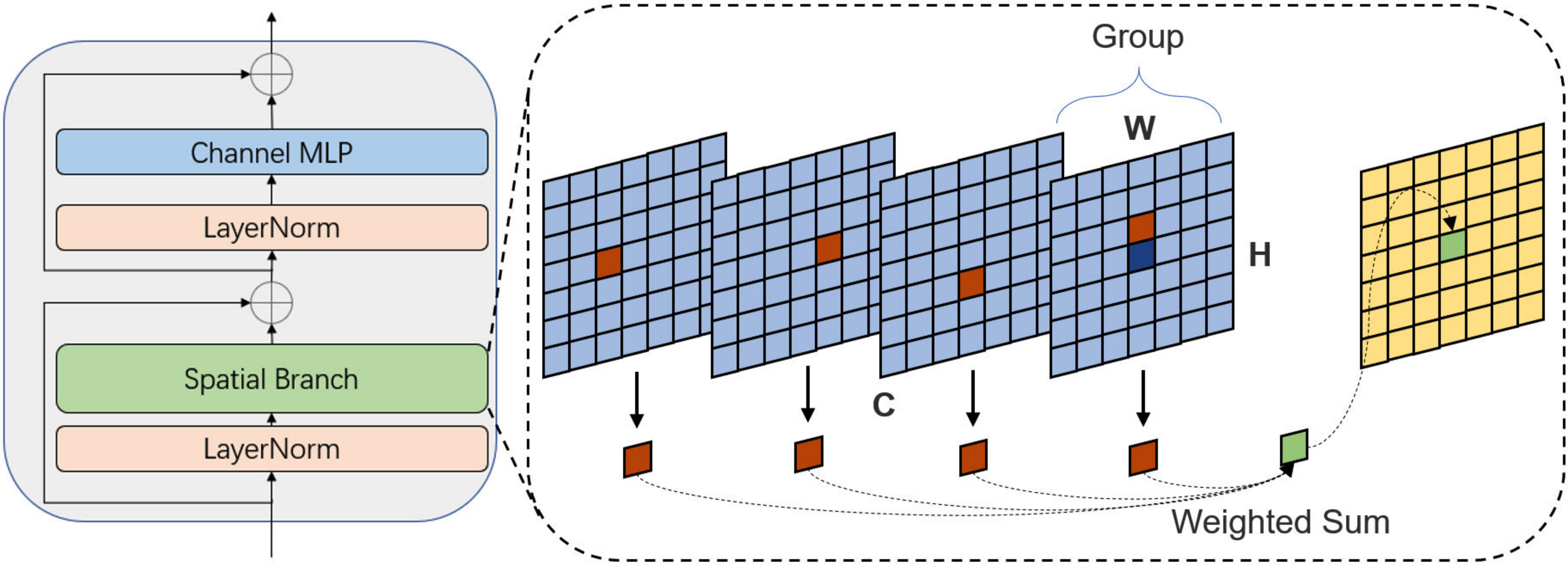}
        \label{fig:channel_only_module_s2mlp}
    }    	 
    \subfigure[S2MLPv2]{   	 		 
        \includegraphics[width=7cm]{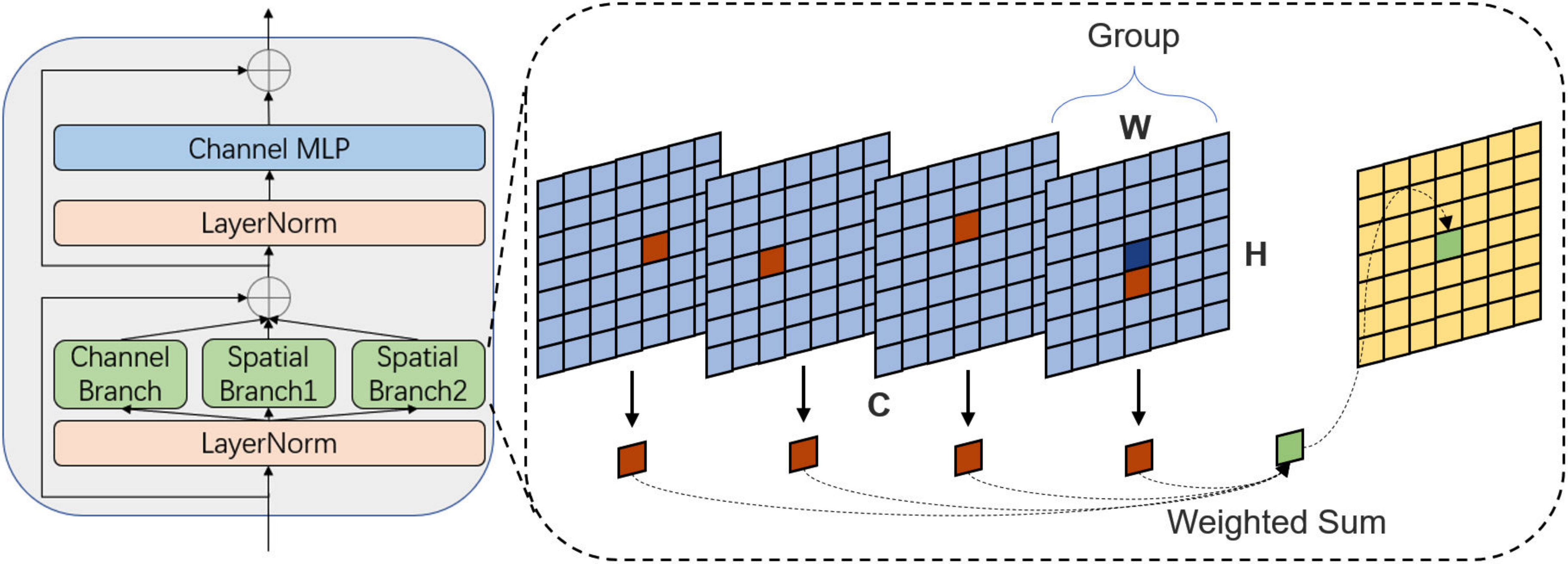}
        \label{fig:channel_only_module_s2mlpv2}
    }
    \subfigure[AS-MLP]{  			 
        \includegraphics[width=7cm]{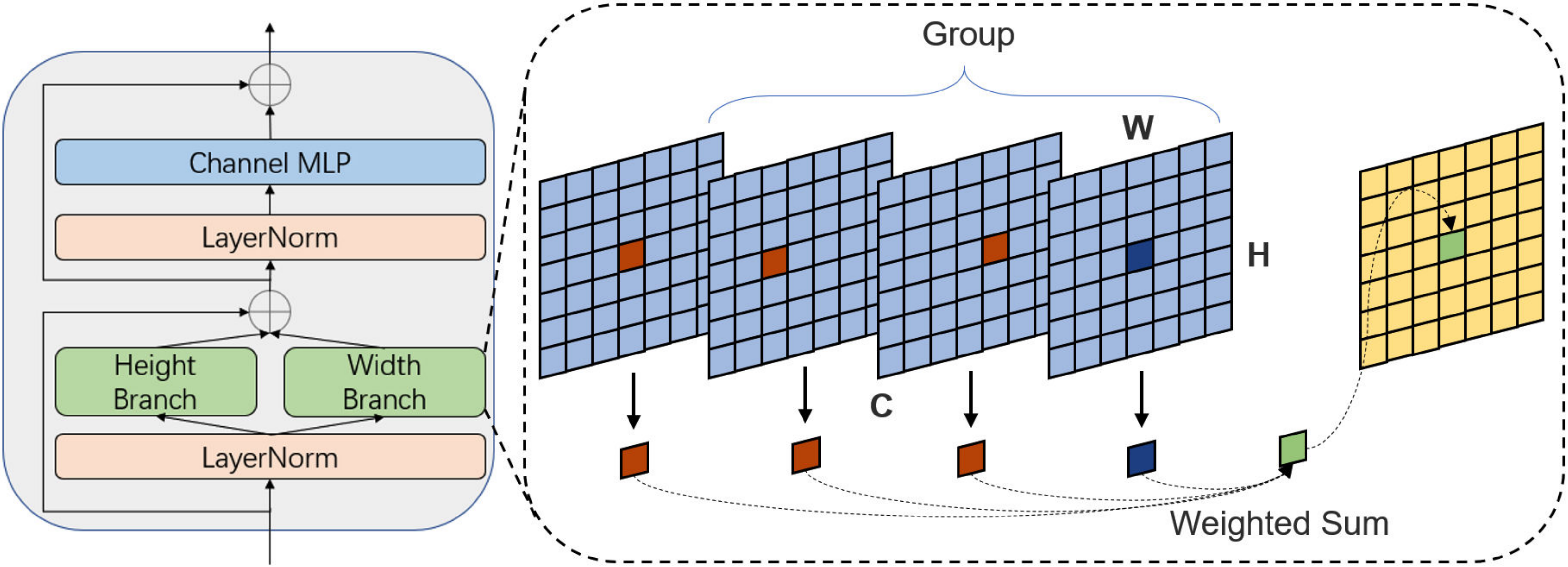}
        \label{fig:channel_only_module_asmlp}
    }    	 
    \subfigure[CycleMLP]{   	 		 
        \includegraphics[width=7cm]{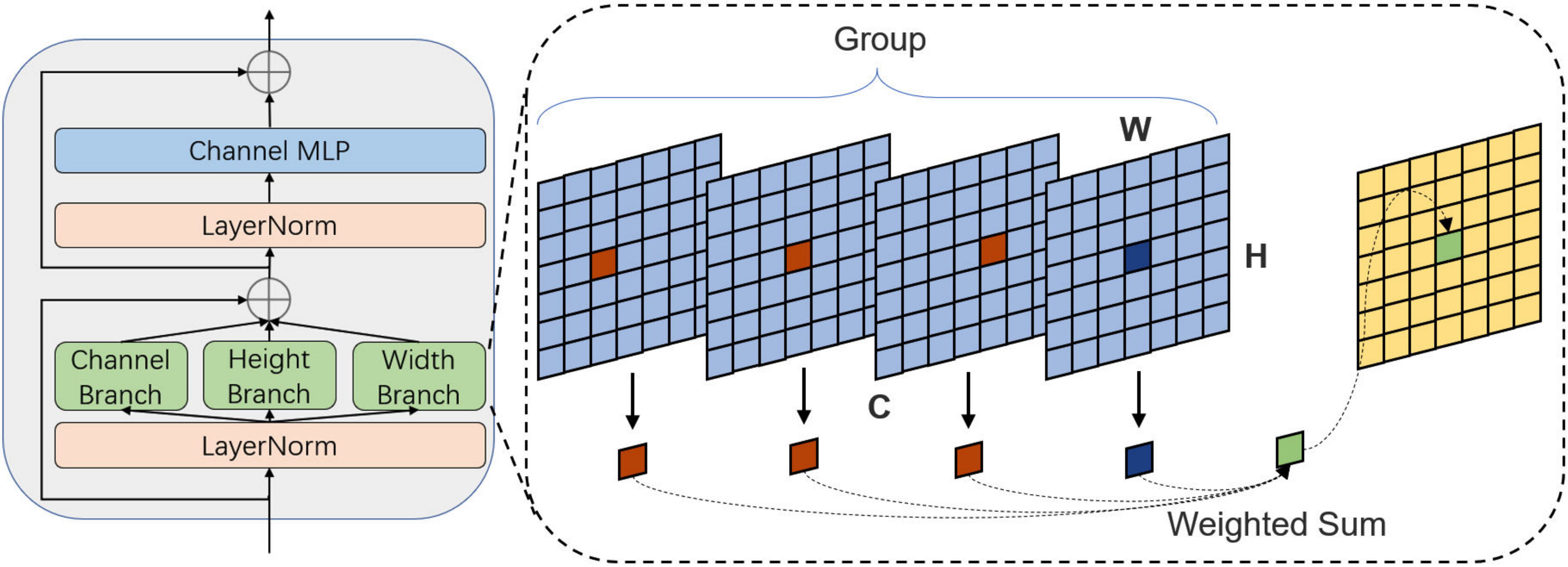}
        \label{fig:channel_only_module_cyclemlp}
    }
    \subfigure[HireMLP]{  			 
        \includegraphics[width=7cm]{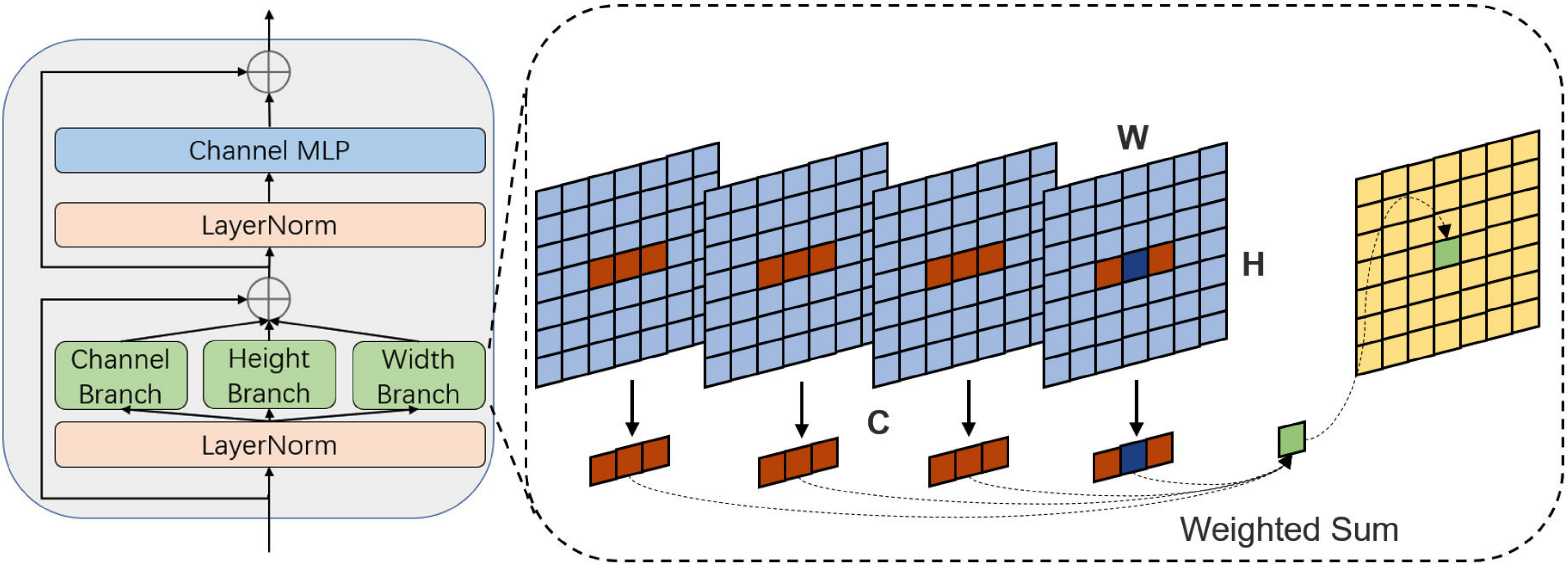}
        \label{fig:channel_only_module_hiremlp}
    }    	 
    \subfigure[MS-MLP]{   	 		 
        \includegraphics[width=7cm]{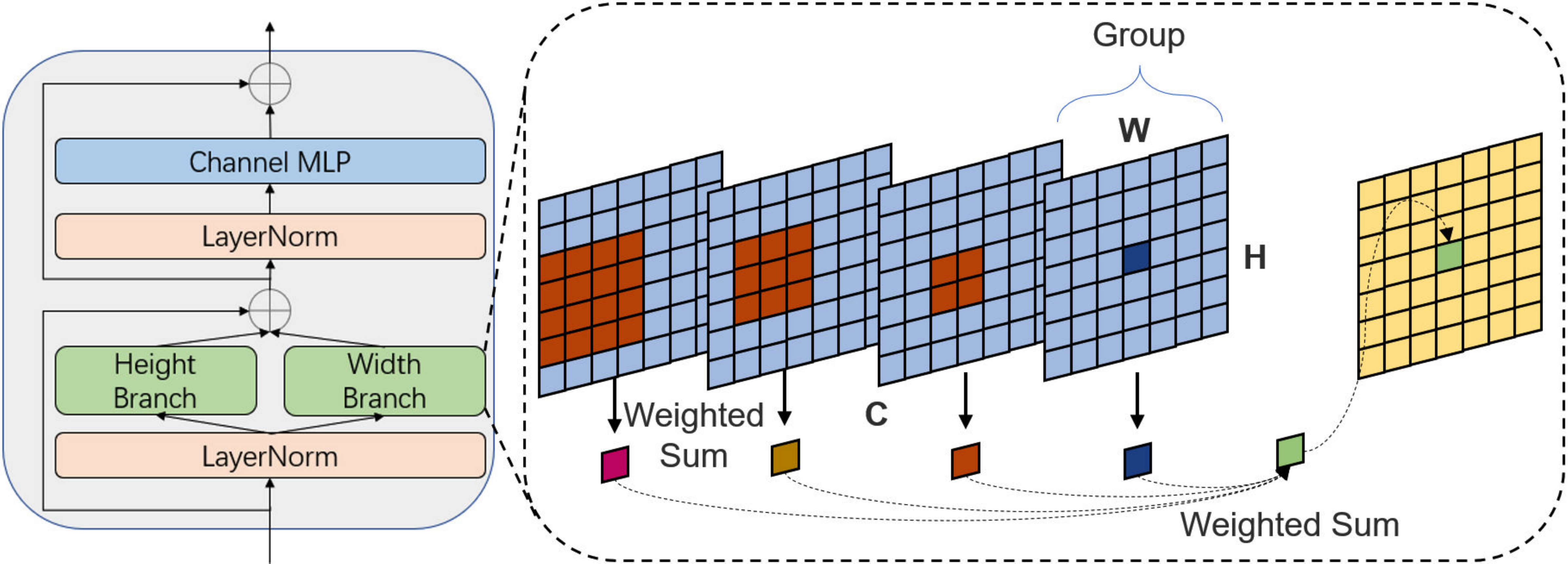}
        \label{fig:channel_only_module_msmlp}
    }
    \subfigure[ActiveMLP]{   	 		 
        \includegraphics[width=11cm]{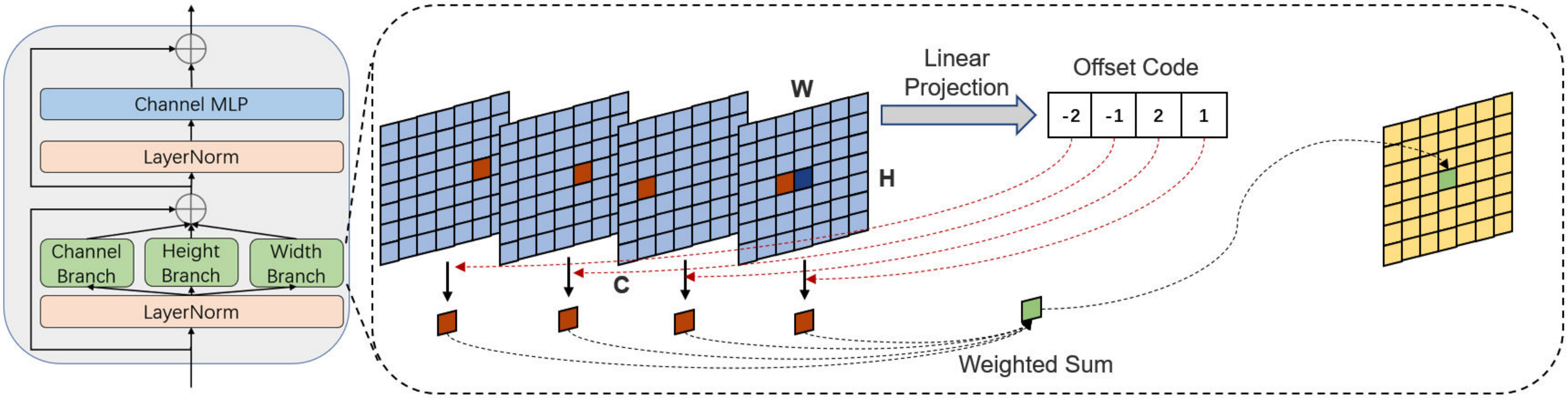}
        \label{fig:channel_only_module_activemlp}
    }
    \caption{Visualizing the module diagram of the MLP-like variants. \textcolor{green}{\emph{Green}} branch is resolution insensitive. The \emph{Channel Branch} and \emph{Linear Projection} both perform a linear projection along the channel dimension. "Group" means that the channel is split into different groups. The expression in the dashed box is consistent with Figure~\ref{fig:paradigmConpare}.}  
    \label{fig:channel_only_module} 
\end{figure}

Yu \emph{et al.}~\cite{yu2021s} propose a spatial-shift MLP-like architecture for vision, called S2MLP. The actual practice is quite simple. As shown in Figure~\ref{fig:channel_only_module_s2mlp}, the proposed spatial-shift module groups $C$ channels into several groups, shifting different channel groups in different directions. The feature map after the shift aligns different token features to the same channel, and then the interaction of the spatial information can be realized after channel projection. Given the simplicity of this approach, Yu \emph{et al.}~\cite{yu2021sv2} exploit the idea of ViP to extend the spatial-shift module into three parallel branches and then fuse the branch features by a split attention module to further improve the network's performance (Figure~\ref{fig:channel_only_module_s2mlpv2}). This newly proposed network is called S2MLPv2.

Unlike grouping and then performing the same shift operation for each group, the Axial Shifted MLP (AS-MLP)~\cite{lian2021mlp} performs different operations within each group that contains a few channels, e.g., every three feature maps in the channel direction are left-shifted, no-shifted, right-shifted, and so on (Figure~\ref{fig:channel_only_module_asmlp}). In addition, AS-MLP uses two parallel branches for horizontal and vertical shifting, where the outputs are added element-wise and projected along the channel dimension for information integration. It is worth mentioning that AS-MLP is also extended with different shifting strategies, allowing the receptive field to be similar to the dilated convolution (atrous convolution)~\cite{yu2015multi,wang2018understanding}.

CycleMLP~\cite{chen2021cyclemlp} was published three days after AS-MLP. Although CycleMLP does not directly shift feature maps, it integrates features at different spatial locations along the channel direction by employing deformable convolution~\cite{dai2017deformable}, an equivalent approach to shifting the feature map. As illustrated in Figure~\ref{fig:channel_only_module_cyclemlp}, CycleMLP and AS-MLP slightly differ as CycleMLP has three branches and AS-MLP has only two branches. Furthermore, CycleMLP relies on split attention to fuse the branched features.

ActiveMLP~\cite{wei2022activemlp} dynamically estimates the offset, rather than manually setting it like AS-MLP and CycleMLP do (Figure~\ref{fig:channel_only_module_activemlp}). It first predicts the spatial locations of helpful contextual features along each direction at the channel level, and then finds and fuses them. This is equivalent to an $1\times1$ deformable convolution~\cite{dai2017deformable}.

HireMLP~\cite{guo2021hire} adopts inner-region and cross-region rearrangements before channel projection to communicate spatial information. The inner-region rearrangement expands the feature map along the height or width direction, and the cross-region rearrangement moves the feature map cyclically along the width or height direction. The HireMLP block still comprises three parallel branches, and the output feature is obtained by adding the branched features (Figure~\ref{fig:channel_only_module_hiremlp}).

The six models mentioned above can communicate only localized information through feature map movement. MS-MLP~\cite{zheng2022mixing} effectively expands the range of receptive fields by mixing neighboring and distant tokens from fine- to coarse-grained levels and then gathering them via a shifting operation. From the implementation aspect, MS-MLP performs depthwise convolution of different sizes before channel alignment (Figure~\ref{fig:channel_only_module_msmlp}). Compared with the global receptive field or local window, there is still some vacancy in the receptive field of MS-MLP.

\textbf{Discussion}: After shifting the feature maps, channel projection is equivalent to sampling features at different locations in different channels for aggregation. In other words, this strategy is an artificially designed deformable convolution. Thus, it may be far better to call these models CNN-like, as only local feature extraction can be performed.

\subsubsection{Spatial and channel projection blocks}

Some variants still retain full space and channel projection. Their module designs are not short of sparkle and enhance performance. Nevertheless, these methods are resolution-sensitive, prohibiting them from being a general vision backbone.

gMLP~\cite{liu2021pay} is the first proposed MLP-Mixer variant. gMLP was developed by Liu \emph{et al.}, who experimented with several design choices for token-mixing architecture and found that spatial projections work well when they are linear and paired with multiplicative gating. In detail, the spatial gating unit first linearly projects the input $X$, $f_{W,b} = WX + b$. Then the output of the spatial gating unit is $X \odot f_{W, b}(X)$, where $\odot$ denotes the element-wise product. The authors found it effective to split $X$ into two independent parts $(X_1, X_2)$ along the channel dimension for the gating function and the multiplicative bypass: $X_{out} = X_1 \odot f_{W, b}(X_2)$. Note that the gMLP block has a channel projection before and after the spatial gating unit. However, there is no channel-mixing MLP anymore. Pleasingly, gMLP achieves good performance in both computer vision and natural language processing tasks.

Lou \emph{et al.}~\cite{lou2021sparse} consider how to scale the MLP-Mixer to more parameters with comparable computational cost, making the model more computationally efficient and better performing. Specifically, they introduce the Mixture-of-Experts (MoE)~\cite{shazeer2017outrageously} scheme into the MLP-Mixer and propose Sparse-MLP(MoE). Carlos \emph{et al.}~\cite{riquelme2021scaling} had already applied MoE on the MLP of the Transformer block, i.e., Channel-mixing MLP in MLP-Mixer, a few months earlier. As a continuation of Carlos' work, Lou expands MoE from Channel-mixing MLP to Token-mixing MLP and achieves some performance gains compared to the primitive MLP-Mixer. More details about MoE can be found in~\cite{shazeer2017outrageously,riquelme2021scaling}.

Yu \emph{et al.}~\cite{yu2021rethinking} impose a circulant-structure constraint on the Token-mixing MLP weights, reducing the spatial projection's sensitivity to spatial translation while preserving the global receptive field. It should be noted that the model is still resolution-sensitive. The authors reduce the number of parameters from $\mathcal{O}(H^2W^2)$ to $\mathcal{O}(HW)$, but do not reduce the computation cost. Therefore, the authors employ a fast Fourier transform to reduce the FLOPs and enhance computational efficiency. 

Finally, and the most ingeniously, Ding \emph{et al.}~\cite{ding2021repmlpnet} propose a novel Structural Re-parameterization technique to merge the convolutional layers into the fully connected layers. Therefore, during training, the proposed RepMLPNet can learn parallel fully connected layers (global receptive field) and convolutional layers (local receptive field) and combine the two via transforming the parameters. Compared with the weight values of the fully connected layers before merging, the re-parameterization resultant weight has larger values around a specific position, suggesting that the model focuses more on the neighborhood. Although large images can be sliced and input to RepMLPNet for feature extraction, the resolution sensitivity makes it not a general vision backbone.

\subsubsection{Others}

ConvMLP~\cite{li2021convmlp} is another special variant, which is lightweight, stage-wise, and comprises a co-design of convolutional layers. ConvMLP replaces the Token-mixing MLP with a $3 \times 3$ depth-wise convolution, and therefore we consider ConvMLP as a pure CNN model rather than an MLP-like model.

In addition, LIT~\cite{pan2021less} replaces the self-attention layers with MLP in the first two stages of a pyramid vision transformer. UniNet~\cite{liu2021uninet} jointly searches the optimal combination of convolution, self-attention, and MLP for building a series of all-operator network architectures with high performances on visual tasks. However, these methods are beyond the scope of Vision MLP.

\subsection{Receptive Field and Complexity Analysis}
The main novelty of MLP is allowing the model to \emph{autonomously learn the global receptive field from raw data}. However, do these so-called MLP-like variants still hold to the original intent? Immediately following the module design, we compare and analyze the receptive field of these blocks to provide a more in-depth presentation for those MLP-like variants. Following that, we compare and analyze the complexity of different modules, with the corresponding results reported in Table~\ref{table:blocks}. A comparison of the block's spatial sensitivity, channel sensitivity, and resolution sensitivity is also provided.

\subsubsection{Receptive field}

\begin{figure}[!htb]
  \begin{minipage}[b]{1.0\linewidth}
    \centering
    \centerline{\includegraphics[width=\linewidth]{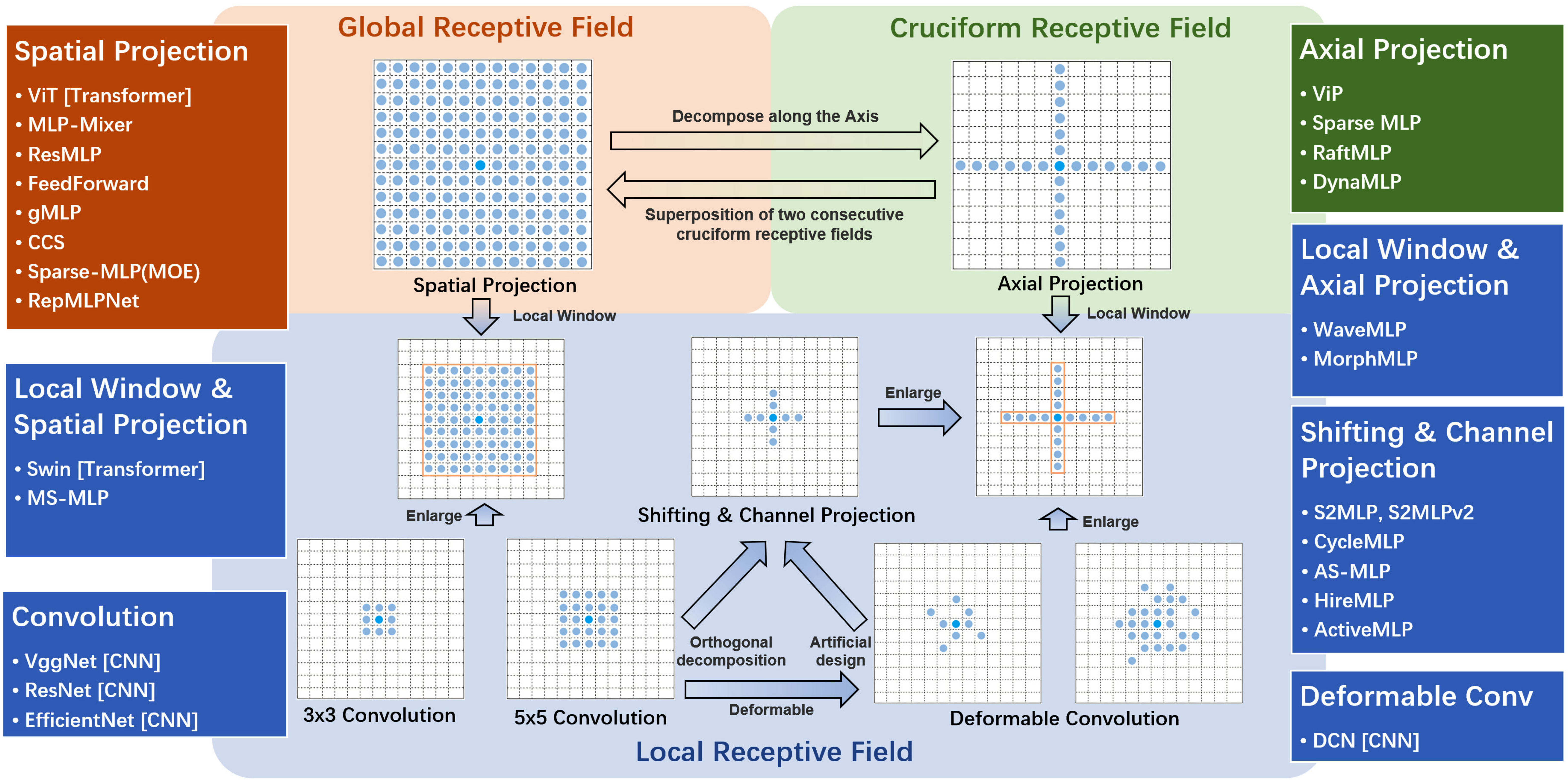}}
  \end{minipage}
  \caption{
    Schematic diagram of different receptive fields, including global, cruciform, and local. The bright blue dot indicates the position of the encoded token, and the light blue dots are other locations involved in the calculation. The range of the blue dots constitutes the receptive field. The orange rectangle is the Local Window, which covers a larger region than the commonly used convolutional kernels. The transition between different receptive fields is marked next to the arrow. The corresponding networks are listed on the left and right sides. Although MS-MLP~\cite{zheng2022mixing} uses Shifting \& Channel Projection, its receptive field is most similar to the Local Window \& Spatial Projection due to the depth-wise convolution before shifting feature maps. The local receptive fields formed by MLP-like variants can be realized by convolution, making them lack an essential difference from CNNs.
  }
  \label{fig:receptivefield}
\end{figure}

\emph{Receptive field} was first used by Sherrington~\cite{sherrington1906observations} in 1906 to describe the skin area that could trigger a scratch reflex in a dog. Nowadays, the term \emph{receptive field} is also used in describing artificial neural networks, which is deemed as the size of the region in the input that produces the output value. It measures the relevance of an output feature to the input region. Different information aggregation methods will generate different receptive fields, which we divide into three categories, global, cruciform, and local. Convolutional neural networks have local receptive fields, while vision transformers own global receptive fields. Swin Transformer\cite{liu2021swin} introduces the concept of Local Window, reducing the global receptive field to a fixed region independent of the image resolution, which is still more extensive than the standard convolutional layer. Figure~\ref{fig:receptivefield} displays the schematic diagrams of different receptive fields and the corresponding MLP-like variants in detail.

Similar to MLP-Mixer~\cite{tolstikhin2021mlp}, the full spacial projection in gMLP~\cite{liu2021pay}, Sparse-MLP(MoE)~\cite{lou2021sparse}, CCS~\cite{yu2021rethinking}, and RepMLPNet~\cite{ding2021repmlpnet} still retain the global receptive field, i.e., the encoded features of each token are the weighted sum of all input token features. It must be acknowledged that this global receptive field is at the patch level, not at the pixel level in the traditional sense. In other words, the globalness is approximated by patch partition, similar to the Transformer-based methods. The size of the patch affects the final result and the network's computational complexity. Notably, the patch partition is a strong artificial assumption that is always ignored.

To balance long-distance dependence and computational cost, axial projection decomposes the full spacial projection orthogonally, i.e., along with horizontal and vertical directions. The projection on both axes is made serially (RaftMLP~\cite{tatsunami2021raftmlp}) or parallel (ViP~\cite{hou2021vision}, Sparse MLP~\cite{tang2021sparse}, DynaMixer~\cite{wang2022dynamixer}). Thus, the token encoded only interacts with horizontal or vertical tokens in a single projection and forms a cruciform receptive field, which retains horizontal and vertical long-range dependence. However, if the token of interest and the current token are not at the same horizontal or vertical area, the two tokens cannot interact. 

The global and cruciform receptive fields are required to cover the entire height and width of the space, resulting in a one-to-one correspondence between the neurons' number in the fully connected layer and the image resolution, further restricting the model to utilize images of a specific size. To eliminate resolution sensitivity, many MLP-like variants (Blue box in Figure~\ref{fig:receptivefield}) choose to use local receptive fields. Mainly, two approaches are adopted: local window and channel mapping after shifting feature maps. However, these operations can be achieved by expanding the convolutional kernel size and using deformable convolution, making these variants not fundamentally different from CNNs. A concern is that these MLP-like variants abandon the global receptive field, a significant MLP feature.

Virtually, the cruciform and local receptive field is a particular case of the global receptive field, e.g., the weight value is approximately $0$ except for a specific area. Therefore, these two receptive field types are equivalent to learning only a small part of the weights of the global receptive field and setting the other weights to be constantly 0. This is also an artificial inductive bias, similar to the locality introduced by the convolutional layer.

\subsubsection{Complexity analysis}
\label{subsec:complexity}

Since all network blocks mentioned above contain the channel-mixing MLP module, where the number of parameters is $\mathcal{O}(C^2)$, and the FLOPs are $\mathcal{O}(HWC^2)$, we ignore this item in the later analysis and focus mainly on the module used for spatial information fusion, i.e., Token-mixing MLP and its variants. Table~\ref{table:blocks} lists the comparison results of the attribute and complexity analysis of different spatial information fusion modules, where the network names are used to name each module for ease understanding. The complexity is referenced from the analysis in the original papers, respectively.

The full spacial projection in the MLP-Mixer~\cite{tolstikhin2021mlp} and its variants contains $\mathcal{O}(H^2W^2)$ parameters and has $\mathcal{O}(H^2W^2C)$ FLOPs, both quadratic with the image resolution. Theoretically, it is difficult to apply the network to large resolution images with the current computational power. Hence, a compromise to enhance computational efficiency significantly increases the patch size, e.g.,$14 \times 14$ or $16 \times 16$, and the information extracted is too coarse to discriminate small objects. The only difference is that CCS~\cite{yu2021rethinking} constructs an $HW \times HW$ weight matrix using weight vectors of length $HW$ and uses a fast Fourier transform to reduce the computational cost.

\begin{table}[]
    \centering
    \caption{Comparison of different spatial information fusion modules. $H$, $W$, and $C$ are the feature map's height, width, and channel numbers, respectively. $L$ is the local window size. \emph{Spatial} refers to whether feature extraction is sensitive to the spatial location of objects, \emph{specific} means true, while \emph{agnostic} means false. \emph{Channel} refers whether weights are shared between channels, \emph{agnostic} shares weights between all channels, \emph{group-specific} shares weights between groups, and \emph{specific} does not share. \emph{Resolution Sensitive} refers to whether the module is resolution sensitive.}
    \resizebox{\textwidth}{!}{
    \begin{tabular}{c|l|ccccc}
        \toprule
        \begin{tabular}[c]{@{}c@{}}Spatial\\ Operation\end{tabular} & Block & Spatial & Channel & \begin{tabular}[c]{@{}c@{}}Resolution\\ Sensitive\end{tabular} & Params & FLOPs \\ 
        \midrule
        \multirow{7}{*}{\begin{tabular}[c]{@{}c@{}}Spatial\\ Projection\end{tabular}} & MLP-Mixer~\cite{tolstikhin2021mlp}  & specific & agnostic & True & $\mathcal{O}(H^2W^2)$ & $\mathcal{O}(H^2W^2C)$ \\
         & ResMLP~\cite{touvron2021resmlp}  & specific & agnostic & True & $\mathcal{O}(H^2W^2)$ & $\mathcal{O}(H^2W^2C)$ \\
         & FeedForward~\cite{melas2021you}  & specific & agnostic & True & $\mathcal{O}(H^2W^2)$ & $\mathcal{O}(H^2W^2C)$ \\
         & gMLP~\cite{liu2021pay}  & specific & agnostic & True & $\mathcal{O}(H^2W^2)$ & $\mathcal{O}(H^2W^2C)$ \\
         & Sparse-MLP(MoE)~\cite{lou2021sparse}  & specific & agnostic & True & $\mathcal{O}(H^2W^2)$ & $\mathcal{O}(H^2W^2C)$ \\
         & CCS~\cite{yu2021rethinking}  & agnostic & group-specific & True & $\mathcal{O}(HW)$ & $\mathcal{O}(HW\log(HW)C)$ \\
         & RepMLPNet~\cite{ding2021repmlpnet}  & specific & group-specific & True & $\mathcal{O}(H^2W^2 + C^2)$ & $\mathcal{O}(H^2W^2C)$ \\\midrule
        \multirow{6}{*}{\begin{tabular}[c]{@{}c@{}}Axial\\ Projection\end{tabular}} & RaftMLP~\cite{tatsunami2021raftmlp}  & specific & agnostic & True & $\mathcal{O}(H^2+W^2)$ & $\mathcal{O}(HWC(H+W))$ \\
         & ViP~\cite{hou2021vision}  & specific & specific & True & $\mathcal{O}(H^2+W^2+C^2)$ & $\mathcal{O}(HWC(H+W+C))$ \\
         & Sparse MLP~\cite{tang2021sparse}  & specific & specific & True & $\mathcal{O}(H^2+W^2+C^2)$ & $\mathcal{O}(HWC(H+W+C))$ \\ 
         & DynaMixer~\cite{wang2022dynamixer}  & specific & specific & True & $\mathcal{O}(H^3+W^3+C^2)$ & $\mathcal{O}(HWC(H^2+W^2+C))$ \\ 
         & WaveMLP~\cite{tang2021image}  & specific & specific & False & $\mathcal{O}(2L^2 + C^2)$ & $\mathcal{O}(HWC(L+L+C))$ \\ 
         & MorphMLP~\cite{zhang2021morphmlp}  & specific & specific & False & $\mathcal{O}(2L^2 + C^2)$ & $\mathcal{O}(HWC(L+L+C))$ \\ 
         \midrule
        \multirow{6}{*}{\begin{tabular}[c]{@{}c@{}}Shifting \& \\ Channel\\ Projection\end{tabular}} & S2MLP~\cite{yu2021s} & agnostic & specific & False & $\mathcal{O}(C^2)$ & $\mathcal{O}(HWC^2)$ \\
         & S2MLPv2~\cite{yu2021sv2} & agnostic & specific & False & $\mathcal{O}(C^2)$ & $\mathcal{O}(HWC^2)$ \\
         & AS-MLP~\cite{lian2021mlp} & agnostic & specific & False & $\mathcal{O}(C^2)$ & $\mathcal{O}(HWC^2)$ \\
         & CycleMLP~\cite{chen2021cyclemlp} & agnostic & specific & False & $\mathcal{O}(C^2)$ & $\mathcal{O}(HWC^2)$ \\
         & HireMLP~\cite{guo2021hire} & agnostic & specific & False & $\mathcal{O}(C^2)$ & $\mathcal{O}(HWC^2)$ \\ 
         & MS-MLP~\cite{zheng2022mixing} & agnostic & specific & False & $\mathcal{O}(C^2)$ & $\mathcal{O}(HWC^2)$ \\
         & ActiveMLP~\cite{wei2022activemlp} & agnostic & specific & False & $\mathcal{O}(C^2)$ & $\mathcal{O}(HWC^2)$ \\
         \bottomrule
    \end{tabular}
    }
    \label{table:blocks}
\end{table}

In contrast and similar to RaftMLP~\cite{tatsunami2021raftmlp}, the axial projection, the orthogonal decomposition of the full spatial projection, reduces the parameter cardinality from $\mathcal{O}(H^2W^2)$ to $\mathcal{O}(H^2+W^2)$, and the FLOPs from $\mathcal{O}(H^2W^2C)$ to $\mathcal{O}(HWC(H+W))$.If three parallel branches are used, the number of parameters and the FLOPs are $\mathcal{O}(H^2+W^2+C^2)$ and $\mathcal{O}(HWC(H+W+C))$, respectively. Notice that fusing information from multiple branches does not increase the computational complexity, such as splitting attention and dimensionality reduction after channel concatenation. If there is a Local Window, the number of parameters and FLOPs are further reduced to $\mathcal{O}(2L^2 + C^2)$ and $\mathcal{O}(HWC(L+L+C))$, where $L$ is the window size, supposing a channel branch exists. DynaMixer~\cite{wang2022dynamixer} requires dynamic estimation of the weight matrix, which leads to higher complexity.

The approach based on shifting the feature map and channel projection further reduce computational complexity, i.e., the number of parameters is $\mathcal{O}(C^2)$ with $\mathcal{O}(HWC^2)$ FLOPs. MS-MLP~\cite{zheng2022mixing} adds some depthwise convolutions, ActiveMLP~\cite{wei2022activemlp} adds some channel projections, but both do not affect the overall complexity. Moreover, the number of weights is decoupled from the image resolution to no longer constrain these variants.

It is worth raising that computational complexity is only one of the determinants of inference time, as reshaping and shifting feature maps is also time-consuming. In addition, reducing complexity does not mean that the proposed network has fewer parameters. Conversely, the various networks retain a comparable number of parameters (Table~\ref{table:ImageNet-MLP}). This allows networks of lower complexity to have more layers and more channels.

\subsection{Discussion}

The bottleneck of Token-mixing MLP (Section~\ref{subsec:mlp-question}) induces researchers to redesign the block. Recently released MLP-like variants reduce the model's computational complexity, dynamic aggregation information, and resolving image resolution sensitivity. Specifically, researchers decompose the full spacial projection orthogonally, restrict interaction within a local window, perform channel projection after shifting feature maps, and make other artificial designs. These careful and clever designs demonstrate that researchers have noticed that the current amount of data and computational power is insufficient for pure MLPs. Comparing the computational complexity has a theoretical significance, but it is not the only determinant of inference time and final model efficiency. Analysis of the receptive field shows that the new paradigm is instead moving towards the old paradigm. To put it more bluntly, the development of MLP heads back to the way of CNNs. Hence, we still need to make efforts to balance long-distance dependence and image resolution sensitivity.

\section{Architecture of MLP Variants}
\label{sec:stages}

\begin{figure}[!htb]
  \begin{minipage}[b]{1.0\linewidth}
    \centering
    \centerline{\includegraphics[width=\linewidth]{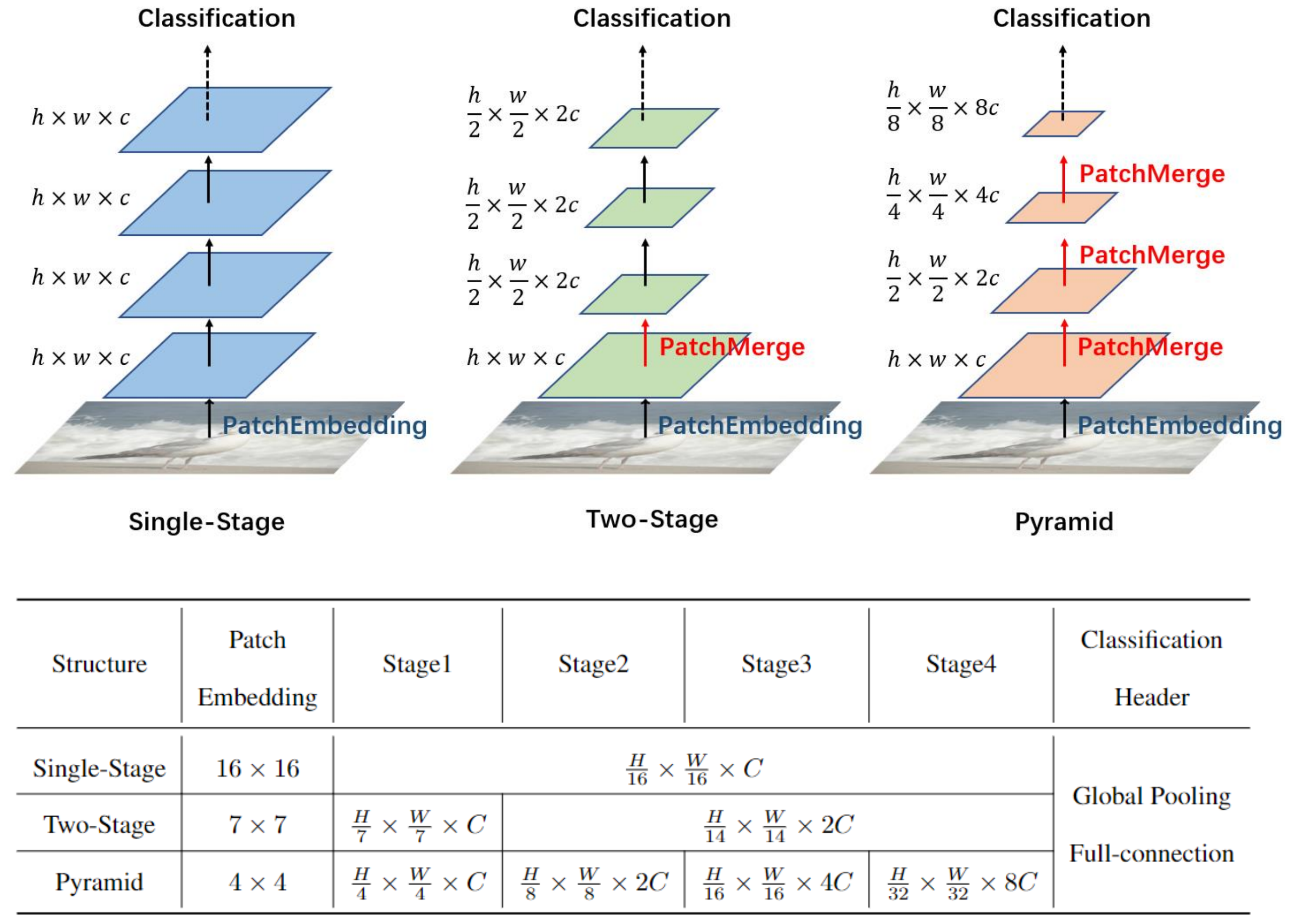}}
  \end{minipage}
  \caption{
     Comparison of different hierarchical architectures for classification models. After patch embedding, the feature map size is $h\times w \times c$, where $h$, $w$, $c$ are the height, width and channel numbers. There is a patch merging operation between every two stages, usually $2 \times 2$ patches are merged, and the number of channels doubles. The resolutions of the feature maps are different, usually $h=H/16$ in Single-Stage, $h=H/7$ in Two-Stage, and $h=H/4$ in Pyramid, where $H$ and $W$ are the height and width of the input image.
  }
  \label{fig:stages}
\end{figure}

Multiple blocks are stacked to form an architecture via selecting a network structure. According to our investigation, the traditional structures for classification models are also applicable to MLP-like architectures, which can be divided into three categories (Figure~\ref{fig:stages}): (1) single-stage structure inherited from the ViT~\cite{dosovitskiy2020image}, (2) two-stage structure with smaller patches in the early stage and larger in the later stage, and (3) CNN-like pyramid structure. And in each stage, there are multiple identical blocks. Table~\ref{tab:framework_classification} illustrates the stacking structures of MLP-Mixer and its variants.

\subsection{From Single-stage to Pyramid}

MLP-Mixer~\cite{tolstikhin2021mlp} inherits the “isotropic” design of ViT~\cite{dosovitskiy2020image}, i.e., after the patch embedding, each block does not change the size of the feature map. This is called a \emph{single-stage} structure. Models of this design include FeedForward~\cite{melas2021you}, ResMLP~\cite{touvron2021resmlp}, gMLP~\cite{liu2021pay}, S2MLP~\cite{yu2021s},  CCS~\cite{yu2021rethinking}, RaftMLP~\cite{tatsunami2021raftmlp}, and Sparse-MLP(MoE)~\cite{lou2021sparse}. Due to the limited computing resources, the patch partition during patch embedding of the single-stage model is usually large, e.g., $16\times16$ or $14\times14$, with the coarse-grained patch partition limiting the subsequent feature fineness. Although the impact is not significant for single-object classification, it impacts  many downstream tasks such as object detection and instance segmentation, especially for small targets.

\begin{table}[]
\centering
\small
\caption{Stacking structures for MLP-Mixer and its variants.}
    \resizebox{\textwidth}{!}{
    \begin{tabular}{@{}c|c|c|c@{}}
    \toprule
    \emph{Spatial Operation} & \textbf{Single-Stage} & \textbf{Two-Stage} & \textbf{Pyramid} \\
    \midrule
     
    \textbf{Spatial Projection} & \makecell[c]{MLP-Mixer~\cite{tolstikhin2021mlp}, ResMLP~\cite{touvron2021resmlp}, \\gMLP~\cite{liu2021pay}, FeedForward~\cite{melas2021you},\\ CCS~\cite{yu2021rethinking}, Sparse-MLP(MoE)~\cite{lou2021sparse}} & -- & RepMLPNet~\cite{ding2021repmlpnet} \\
    \midrule
    
    \textbf{Axial Projection} & RaftMLP~\cite{tatsunami2021raftmlp}  & \makecell[c]{ViP~\cite{hou2021vision},\\ DynaMixer~\cite{wang2022dynamixer}} & \makecell[c]{Sparse MLP~\cite{tang2021sparse}, \\ WaveMLP~\cite{tang2021image}, \\ MorphMLP~\cite{zhang2021morphmlp}} \\
    \midrule
    
    \textbf{\makecell[c]{Shifting \& \\ Channel Projection}}  & S2MLP~\cite{yu2021s}  & S2MLPv2~\cite{yu2021sv2} & \makecell[c]{CycleMLP~\cite{chen2021cyclemlp}, AS-MLP~\cite{lian2021mlp}, \\ HireMLP~\cite{guo2021hire}, MS-MLP~\cite{zheng2022mixing}, \\
    ActiveMLP~\cite{wei2022activemlp}} \\
    
    \bottomrule
    \end{tabular}
    }
\label{tab:framework_classification}
\end{table}

Intuitively, smaller patches are beneficial in modeling fine-grained details in the images and tend to achieve higher recognition accuracy. Vision Permutator~\cite{hou2021vision} further proposes a \emph{two-staged} configuration. Specifically, the network considers $7 \times 7$ patch slices in the initial patch embedding and performs a $2 \times 2$ patch merge after a few layers. During patch merging, the height and width of the feature map halve while the channels double. Comparing with the $14 \times 14$ patch embedding, encoding fine-level patch representations brings a slight Top-1 accuracy improvement on ImageNet-1k (from 80.6\% to 81.5\%). S2MLPv2~\cite{yu2021sv2} follows Vision Permutator and achieves a similar Top-1 accuracy improvement on ImageNet-1k (from 80.9\% to 82.0\%), while DynaMixer~\cite{wang2022dynamixer} also adopts two-staged configurations.

If the initial patch size is further reduced, e.g., $4 \times 4$, more patch merging is required subsequently to reduce the number of patches (or tokens), promoting the network to adopt a pyramid structure. Specifically, the entire structure contains four stages, where the feature resolution reduces from $H/4 \times W/4$ to $H/32 \times W/32$, and the output dimension increases accordingly.
Almost all the recently proposed MLP-like variants adopt the pyramid structure (right side of Table~\ref{tab:framework_classification}). Worth mentioning, a convolutional layer can equivalently achieve patch embedding if its kernel size and stride are equal to the patch size. In the pyramid structure, the researchers find that using an overlapping patch embedding provides better results, that is, convolution of $7 \times 7$ with $\text{stride} = 4$ instead of $4 \times 4$, which is similar to ResNet\footnote{The initial embedding layers of ResNet is a $7 \times 7$ convolutional layer with $\text{stride} = 2$ followed by a $2 \times 2$ max-pooling layer.}~\cite{he2016deep,chen2021cyclemlp}.

\subsection{Discussion}

The architecture of MLP-like variants gradually evolves from single-stage to pyramid, with smaller and smaller patch size and higher feature fineness. We believe this development trend is not only to cater to downstream task frameworks such as FPN~\cite{lin2017feature} but also to balance the original intention and computing power. When the patch size decreases, the number of tokens increases accordingly, so token interactions are limited to a small range to control the amount of computation. With a small range of token interactions, the global receptive field can only be preserved by reducing the size of the feature map, and the pyramid structure appears. This is consistent with the CNN concept, where the alternating use of convolutional and pooling layers has been around since 1979~\cite{fukushima1979neural}!

We would like to point out that, it is unfair to compare single-stage and pyramid models directly based on current configurations (the bottom half of Figure~\ref{fig:stages}). What if a $4\times 4$ patch partition is used in a single-stage model? Will it be worse than the pyramid model? Currently, this is unknown. What is known is that the cost of calculations will increase significantly and is constrained by the current computing devices.

\section{Applications of MLP Variants}
\label{sec:performance}

This section reviews the applications of MLP-like variants in computer vision, including image classification, object detection and semantic segmentation, low-level vision, video analysis and point cloud. Due to the short development time of MLP, we focus on the first two aspects and give an intuitive comparison of MLP, CNN, and Transformer-based models. We are limited to some brief introduction for the latter three aspects, as only a few works are currently available.

\subsection{Image Classification}

ImageNet~\cite{deng2009imagenet} is a large vision dataset designed for visual object classification. Since its release, it has been used as a benchmark for evaluating models in computer vision. Classification performance on ImageNet is often regarded as a reflection of the network's ability to extract visual features. After training on ImageNet, the model can be well transferred to other datasets and downstream tasks, e.g., object detection and segmentation, where the transferred part is usually called \emph{vision backbone}.

Table~\ref{table:ImageNet-MLP} compares the performance of current vision MLP models on ImageNet-1k, including Top-1 accuracy, parameters, and FLOPs, where all results are derived from the cited papers. We further divide the MLP models into three configuration types based on the number of parameters, and the rows are sorted by Top-1 accuracy. The results highlight that under the same training configuration, the recently proposed variants bring good performance gains. Table~\ref{table:ImageNet} provides more detailed and comprehensive information. Compared with the latest CNN and Transformer models, MLP-like variants still pose a performance gap. Without the support of extra training data, both CNN and Transformer exceed 87\% Top-1 accuracy, while the MLP-like variant currently achieves only 84.3\%. High performance may benefit from better architecture-specific training strategies, e.g., PeCo~\cite{dong2021peco}, but we do not yet have a training mode specific to MLP. The gap between MLP and other networks is further widened with additional data support.

\begin{table}[]
    \renewcommand\arraystretch{0.55}
    \centering
    \caption{
        Image classification results of MLP-like models on ImageNet-1K benchmark without extra data. The training and testing size is $224\times224$. \emph{Date} means the initial release date on arXiv, where "2021.05" denotes “May, 2021”. \emph{Open Source Code} refers to whether there is officially open-source code.
    }
    \resizebox{\textwidth}{!}{
    \begin{threeparttable}
    \begin{tabular}{ccccccc}
    \toprule
    \multicolumn{1}{c|}{Model}           & \multicolumn{1}{c|}{Date}    & \multicolumn{1}{c|}{Structure}    & \multicolumn{1}{c|}{Top-1 (\%)} & \multicolumn{1}{c|}{Params (M)} & \multicolumn{1}{c|}{FLOPs (G)} & \begin{tabular}[c]{@{}c@{}}Open\\ Source Code\end{tabular} \\ \midrule
    \multicolumn{7}{c}{Small Models}                                                       \\ \midrule
    \multicolumn{1}{c|}{Sparse-MLP(MoE)-S~\cite{lou2021sparse}}      & \multicolumn{1}{c|}{2021.09} & \multicolumn{1}{c|}{Single-stage} & \multicolumn{1}{c|}{71.3}       & \multicolumn{1}{c|}{21}         & \multicolumn{1}{c|}{-}        & False                               
    \\
    \multicolumn{1}{c|}{RepMLPNet-T224~\cite{ding2021repmlpnet}}         & \multicolumn{1}{c|}{2021.12} & \multicolumn{1}{c|}{Pyramid}  & \multicolumn{1}{c|}{76.4}       & \multicolumn{1}{c|}{15.2}         & \multicolumn{1}{c|}{2.8}      & True                                                  \\
    \multicolumn{1}{c|}{ResMLP-12~\cite{touvron2021resmlp}}         & \multicolumn{1}{c|}{2021.05} & \multicolumn{1}{c|}{Single-stage} & \multicolumn{1}{c|}{76.6}       & \multicolumn{1}{c|}{15}         & \multicolumn{1}{c|}{3.0}      & True                                                  \\
    \multicolumn{1}{c|}{Hire-MLP-Ti~\cite{guo2021hire}}       & \multicolumn{1}{c|}{2021.08} & \multicolumn{1}{c|}{Pyramid}      & \multicolumn{1}{c|}{78.9}       & \multicolumn{1}{c|}{17}         & \multicolumn{1}{c|}{2.1}      & False\tnote{b}                                         \\
    \multicolumn{1}{c|}{gMLP-S~\cite{liu2021pay}}            & \multicolumn{1}{c|}{2021.05} & \multicolumn{1}{c|}{Single-stage} & \multicolumn{1}{c|}{79.4}       & \multicolumn{1}{c|}{20}         & \multicolumn{1}{c|}{4.5}      & True                 \\
    \multicolumn{1}{c|}{AS-MLP-T~\cite{lian2021mlp}}          & \multicolumn{1}{c|}{2021.07} & \multicolumn{1}{c|}{Pyramid}      & \multicolumn{1}{c|}{81.3}       & \multicolumn{1}{c|}{28}         & \multicolumn{1}{c|}{4.4}      & True                                                  \\
    \multicolumn{1}{c|}{ViP-Small/7~\cite{hou2021vision}}       & \multicolumn{1}{c|}{2021.06} & \multicolumn{1}{c|}{Two-stage}    & \multicolumn{1}{c|}{81.5}       & \multicolumn{1}{c|}{25}         & \multicolumn{1}{c|}{6.9}      & True                                                  \\
    \multicolumn{1}{c|}{CycleMLP-B2~\cite{chen2021cyclemlp}}       & \multicolumn{1}{c|}{2021.07} & \multicolumn{1}{c|}{Pyramid}      & \multicolumn{1}{c|}{81.6}       & \multicolumn{1}{c|}{27}         & \multicolumn{1}{c|}{3.9}      & True                                                  \\
    \multicolumn{1}{c|}{MorphMLP-T~\cite{zhang2021morphmlp}}       & \multicolumn{1}{c|}{2021.11} & \multicolumn{1}{c|}{Pyramid}    & \multicolumn{1}{c|}{81.6}       & \multicolumn{1}{c|}{23}        & \multicolumn{1}{c|}{3.9}      & False                                                  \\
    \multicolumn{1}{c|}{Sparse MLP-T~\cite{tang2021sparse}}         & \multicolumn{1}{c|}{2021.09} & \multicolumn{1}{c|}{Pyramid}      & \multicolumn{1}{c|}{81.9}       & \multicolumn{1}{c|}{24.1}       & \multicolumn{1}{c|}{5.0}      & False               
    \\
    \multicolumn{1}{c|}{ActiveMLP-T~\cite{wei2022activemlp}}         & \multicolumn{1}{c|}{2022.03} & \multicolumn{1}{c|}{Pyramid}      & \multicolumn{1}{c|}{82.0}       & \multicolumn{1}{c|}{27}       & \multicolumn{1}{c|}{4.0}      & False                
    \\
    \multicolumn{1}{c|}{S2-MLPv2-Small/7~\cite{yu2021sv2}}  & \multicolumn{1}{c|}{2021.08} & \multicolumn{1}{c|}{Two-stage}    & \multicolumn{1}{c|}{82.0}       & \multicolumn{1}{c|}{25}         & \multicolumn{1}{c|}{6.9}      & False                \\
    \multicolumn{1}{c|}{MS-MLP-T~\cite{zheng2022mixing}}         & \multicolumn{1}{c|}{2022.02} & \multicolumn{1}{c|}{Pyramid}      & \multicolumn{1}{c|}{82.1}       & \multicolumn{1}{c|}{28}       & \multicolumn{1}{c|}{4.9}      & True                
    \\
    \multicolumn{1}{c|}{WaveMLP-S~\cite{tang2021image}}         & \multicolumn{1}{c|}{2021.11} & \multicolumn{1}{c|}{Pyramid}      & \multicolumn{1}{c|}{82.6}       & \multicolumn{1}{c|}{30.0}       & \multicolumn{1}{c|}{4.5}      & False\tnote{b}                
    \\
    \multicolumn{1}{c|}{DynaMixer-S~\cite{wang2022dynamixer}}         & \multicolumn{1}{c|}{2022.01} & \multicolumn{1}{c|}{Two-stage}      & \multicolumn{1}{c|}{\textbf{82.7}}       & \multicolumn{1}{c|}{26}       & \multicolumn{1}{c|}{7.3}      & False                
    \\
    \midrule \multicolumn{7}{c}{Medium Models}                                          
    \\ \midrule
    \multicolumn{1}{c|}{FeedForward~\cite{melas2021you}}                & \multicolumn{1}{c|}{2021.05} & \multicolumn{1}{c|}{Single-stage} & \multicolumn{1}{c|}{74.9}       & \multicolumn{1}{c|}{62}         & \multicolumn{1}{c|}{11.4}     & True                                                  \\
    \multicolumn{1}{c|}{Mixer-B/16~\cite{tolstikhin2021mlp}}        & \multicolumn{1}{c|}{2021.05} & \multicolumn{1}{c|}{Single-stage} & \multicolumn{1}{c|}{76.4}       & \multicolumn{1}{c|}{59}         & \multicolumn{1}{c|}{11.7}     & True                                                  \\
    \multicolumn{1}{c|}{Sparse-MLP(MoE)-B~\cite{lou2021sparse}}      & \multicolumn{1}{c|}{2021.09} & \multicolumn{1}{c|}{Single-stage} & \multicolumn{1}{c|}{77.9}       & \multicolumn{1}{c|}{69}         & \multicolumn{1}{c|}{-}        & False                                                 \\
    \multicolumn{1}{c|}{RaftMLP-12~\cite{tatsunami2021raftmlp}}        & \multicolumn{1}{c|}{2021.08} & \multicolumn{1}{c|}{Single-stage}      & \multicolumn{1}{c|}{78.0}       & \multicolumn{1}{c|}{58}         & \multicolumn{1}{c|}{12.0}     & False                                                 \\
    \multicolumn{1}{c|}{ResMLP-36~\cite{touvron2021resmlp}}         & \multicolumn{1}{c|}{2021.05} & \multicolumn{1}{c|}{Single-stage} & \multicolumn{1}{c|}{79.7}       & \multicolumn{1}{c|}{45}         & \multicolumn{1}{c|}{8.9}      & True                                                  \\
    \multicolumn{1}{c|}{Mixer-B/16 + CCS~\cite{yu2021rethinking}}   & \multicolumn{1}{c|}{2021.06} & \multicolumn{1}{c|}{Single-stage} & \multicolumn{1}{c|}{79.8}       & \multicolumn{1}{c|}{57}         & \multicolumn{1}{c|}{11}       & False                                                 \\
    \multicolumn{1}{c|}{RepMLPNet-B224~\cite{ding2021repmlpnet}}         & \multicolumn{1}{c|}{2021.12} & \multicolumn{1}{c|}{Pyramid}  & \multicolumn{1}{c|}{80.1}       & \multicolumn{1}{c|}{68.2}         & \multicolumn{1}{c|}{6.7}      & True                                                  \\
    \multicolumn{1}{c|}{S2-MLP-deep~\cite{yu2021s}}       & \multicolumn{1}{c|}{2021.06} & \multicolumn{1}{c|}{Single-stage} & \multicolumn{1}{c|}{80.7}       & \multicolumn{1}{c|}{51}         & \multicolumn{1}{c|}{9.7}      & False                \\
    \multicolumn{1}{c|}{ViP-Medium/7~\cite{hou2021vision}}      & \multicolumn{1}{c|}{2021.06} & \multicolumn{1}{c|}{Two-stage}    & \multicolumn{1}{c|}{82.7}       & \multicolumn{1}{c|}{55}         & \multicolumn{1}{c|}{16.3}     & True                                                  \\
    \multicolumn{1}{c|}{CycleMLP-B4~\cite{chen2021cyclemlp}}       & \multicolumn{1}{c|}{2021.07} & \multicolumn{1}{c|}{Pyramid}      & \multicolumn{1}{c|}{83.0}       & \multicolumn{1}{c|}{52}         & \multicolumn{1}{c|}{10.1}     & True                                                  \\
    \multicolumn{1}{c|}{AS-MLP-S~\cite{lian2021mlp}}          & \multicolumn{1}{c|}{2021.07} & \multicolumn{1}{c|}{Pyramid}      & \multicolumn{1}{c|}{83.1}       & \multicolumn{1}{c|}{50}         & \multicolumn{1}{c|}{8.5}      & True                                                  \\
    \multicolumn{1}{c|}{Hire-MLP-B~\cite{guo2021hire}}        & \multicolumn{1}{c|}{2021.08} & \multicolumn{1}{c|}{Pyramid}      & \multicolumn{1}{c|}{83.1}       & \multicolumn{1}{c|}{58}         & \multicolumn{1}{c|}{8.1}      & False\tnote{b}                                                 \\
    \multicolumn{1}{c|}{MorphMLP-B~\cite{zhang2021morphmlp}}       & \multicolumn{1}{c|}{2021.11} & \multicolumn{1}{c|}{Pyramid}    & \multicolumn{1}{c|}{83.2}       & \multicolumn{1}{c|}{58}        & \multicolumn{1}{c|}{10.2}      & False                                                  \\
    \multicolumn{1}{c|}{Sparse MLP-B~\cite{tang2021sparse}}         & \multicolumn{1}{c|}{2021.09} & \multicolumn{1}{c|}{Pyramid}      & \multicolumn{1}{c|}{83.4}       & \multicolumn{1}{c|}{65.9}       & \multicolumn{1}{c|}{14.0}     & False                                                 \\ 
    \multicolumn{1}{c|}{MS-MLP-S~\cite{zheng2022mixing}}         & \multicolumn{1}{c|}{2022.02} & \multicolumn{1}{c|}{Pyramid}      & \multicolumn{1}{c|}{83.4}       & \multicolumn{1}{c|}{50}       & \multicolumn{1}{c|}{9.0}      & True                
    \\
    \multicolumn{1}{c|}{ActiveMLP-B~\cite{wei2022activemlp}}         & \multicolumn{1}{c|}{2022.03} & \multicolumn{1}{c|}{Pyramid}      & \multicolumn{1}{c|}{83.5}       & \multicolumn{1}{c|}{52}       & \multicolumn{1}{c|}{10.1}      & False                
    \\
    \multicolumn{1}{c|}{S2-MLPv2-Medium/7~\cite{yu2021sv2}} & \multicolumn{1}{c|}{2021.08} & \multicolumn{1}{c|}{Two-stage}    & \multicolumn{1}{c|}{83.6}       & \multicolumn{1}{c|}{55}         & \multicolumn{1}{c|}{16.3}     & False                \\
    \multicolumn{1}{c|}{WaveMLP-B~\cite{tang2021image}}         & \multicolumn{1}{c|}{2021.11} & \multicolumn{1}{c|}{Pyramid}      & \multicolumn{1}{c|}{83.6}       & \multicolumn{1}{c|}{63.0}       & \multicolumn{1}{c|}{10.2}      & False\tnote{b}                
    \\
    \multicolumn{1}{c|}{DynaMixer-M~\cite{wang2022dynamixer}}         & \multicolumn{1}{c|}{2022.01} & \multicolumn{1}{c|}{Two-stage}      & \multicolumn{1}{c|}{\textbf{83.7}}       & \multicolumn{1}{c|}{57}       & \multicolumn{1}{c|}{17.0}      & False                
    \\
    \midrule \multicolumn{7}{c}{Large Models}                                          
    \\ \midrule
    \multicolumn{1}{c|}{Sparse-MLP(MoE)-L~\cite{lou2021sparse}}      & \multicolumn{1}{c|}{2021.09} & \multicolumn{1}{c|}{Single-stage} & \multicolumn{1}{c|}{79.2}       & \multicolumn{1}{c|}{130}        & \multicolumn{1}{c|}{-}        & False                                                 \\
    \multicolumn{1}{c|}{S2-MLP-wide~\cite{yu2021s}}       & \multicolumn{1}{c|}{2021.06} & \multicolumn{1}{c|}{Single-stage} & \multicolumn{1}{c|}{80.0}       & \multicolumn{1}{c|}{71}         & \multicolumn{1}{c|}{14.0}     & False                \\
    \multicolumn{1}{c|}{gMLP-B~\cite{liu2021pay}}            & \multicolumn{1}{c|}{2021.05} & \multicolumn{1}{c|}{Single-stage} & \multicolumn{1}{c|}{81.6}       & \multicolumn{1}{c|}{73}         & \multicolumn{1}{c|}{15.8}     & True                 \\
    \multicolumn{1}{c|}{RepMLPNet-L256\tnote{a}~\cite{ding2021repmlpnet}}         & \multicolumn{1}{c|}{2021.12} & \multicolumn{1}{c|}{Pyramid}  & \multicolumn{1}{c|}{81.8}       & \multicolumn{1}{c|}{117.7}         & \multicolumn{1}{c|}{11.5}      & True                                                  \\
    \multicolumn{1}{c|}{ViP-Large/7~\cite{hou2021vision}}       & \multicolumn{1}{c|}{2021.06} & \multicolumn{1}{c|}{Two-stage}    & \multicolumn{1}{c|}{83.2}       & \multicolumn{1}{c|}{88}         & \multicolumn{1}{c|}{24.3}     & True                                                  \\
    \multicolumn{1}{c|}{CycleMLP-B5~\cite{chen2021cyclemlp}}       & \multicolumn{1}{c|}{2021.07} & \multicolumn{1}{c|}{Pyramid}      & \multicolumn{1}{c|}{83.2}       & \multicolumn{1}{c|}{76}         & \multicolumn{1}{c|}{12.3}     & True                                                  \\
    \multicolumn{1}{c|}{AS-MLP-B~\cite{lian2021mlp}}          & \multicolumn{1}{c|}{2021.07} & \multicolumn{1}{c|}{Pyramid}      & \multicolumn{1}{c|}{83.3}       & \multicolumn{1}{c|}{88}         & \multicolumn{1}{c|}{15.2}     & True                                                  \\
    \multicolumn{1}{c|}{Hire-MLP-L~\cite{guo2021hire}}        & \multicolumn{1}{c|}{2021.08} & \multicolumn{1}{c|}{Pyramid}      & \multicolumn{1}{c|}{83.4}       & \multicolumn{1}{c|}{96}         & \multicolumn{1}{c|}{13.5}     & False\tnote{b}                                         
    \\
    \multicolumn{1}{c|}{MorphMLP-L~\cite{zhang2021morphmlp}}       & \multicolumn{1}{c|}{2021.11} & \multicolumn{1}{c|}{Pyramid}    & \multicolumn{1}{c|}{83.4}       & \multicolumn{1}{c|}{76}        & \multicolumn{1}{c|}{12.5}      & False                                                 
    \\
    \multicolumn{1}{c|}{ActiveMLP-L~\cite{wei2022activemlp}}         & \multicolumn{1}{c|}{2022.03} & \multicolumn{1}{c|}{Pyramid}      & \multicolumn{1}{c|}{83.6}       & \multicolumn{1}{c|}{76}       & \multicolumn{1}{c|}{12.3}      & False                
    \\
    \multicolumn{1}{c|}{MS-MLP-B~\cite{zheng2022mixing}}         & \multicolumn{1}{c|}{2022.02} & \multicolumn{1}{c|}{Pyramid}      & \multicolumn{1}{c|}{83.8}       & \multicolumn{1}{c|}{88}       & \multicolumn{1}{c|}{16.1}      & True                
    \\
    \multicolumn{1}{c|}{DynaMixer-L~\cite{wang2022dynamixer}}         & \multicolumn{1}{c|}{2022.01} & \multicolumn{1}{c|}{Two-stage}      & \multicolumn{1}{c|}{\textbf{84.3}}       & \multicolumn{1}{c|}{97}       & \multicolumn{1}{c|}{27.4}      & False                
    \\
    \bottomrule
    \end{tabular}
    \begin{tablenotes}
        \footnotesize 
        \item[a] The training and testing size is $256\times256$.
        \item[b] Unofficial code and weights are open sourced at \url{https://github.com/sithu31296/image-classification}.
    \end{tablenotes} 
    \end{threeparttable}
    }
    \label{table:ImageNet-MLP}
\end{table}

\begin{table}[]
\renewcommand\arraystretch{0.65}
    \centering
    \caption{
        Image classification results of representative CNN, vision transformer, and MLP-like models on ImageNet-1K benchmark. \emph{Pre-trained Dataset} column provides extra data information. PloyLoss, PeCo and Meta Pseudo Labels are different training strategies, where the used model is in the bracket.
    }
    \resizebox{\textwidth}{!}{
    \begin{tabular}{ccccc}
    \toprule
    \multicolumn{1}{c|}{Model}       & \multicolumn{1}{c|}{Pre-trained Dataset}  & \multicolumn{1}{c|}{Top-1 (\%)} & \multicolumn{1}{c|}{Params (M)} & \multicolumn{1}{c}{FLOPs (G)} \\ \midrule
    \multicolumn{5}{c}{CNN-based}                                                       
    \\ \midrule
    \multicolumn{1}{c|}{VGG-16~\cite{simonyan2014very}}         & \multicolumn{1}{c|}{--}          & \multicolumn{1}{c|}{71.5}       & \multicolumn{1}{c|}{134}       & \multicolumn{1}{c}{15.5}             
    \\
     \multicolumn{1}{c|}{Xception~\cite{chollet2017xception}}         & \multicolumn{1}{c|}{--}          & \multicolumn{1}{c|}{79.0}       & \multicolumn{1}{c|}{22.9}       & \multicolumn{1}{c}{--}             
    \\
    \multicolumn{1}{c|}{Inception-ResNet-V2~\cite{szegedy2017inception}}         & \multicolumn{1}{c|}{--}          & \multicolumn{1}{c|}{80.1}       & \multicolumn{1}{c|}{--}       & \multicolumn{1}{c}{--}          
    \\
    \multicolumn{1}{c|}{ResNet-50~\cite{he2016deep,wightman2021resnet}}         & \multicolumn{1}{c|}{--}          & \multicolumn{1}{c|}{80.4}       & \multicolumn{1}{c|}{25.6}       & \multicolumn{1}{c}{4.1}         
    \\
    \multicolumn{1}{c|}{ResNet-152~\cite{he2016deep,wightman2021resnet}}        & \multicolumn{1}{c|}{--}         & \multicolumn{1}{c|}{82.0}       & \multicolumn{1}{c|}{60.2}       & \multicolumn{1}{c}{11.5}                             
    \\
    \multicolumn{1}{c|}{RegNetY-8GF~\cite{radosavovic2020designing,wightman2021resnet}}    & \multicolumn{1}{c|}{--}          & \multicolumn{1}{c|}{82.2}       & \multicolumn{1}{c|}{39}         & \multicolumn{1}{c}{8.0}                              
    \\
    \multicolumn{1}{c|}{RegNetY-16GF~\cite{radosavovic2020designing}}    & \multicolumn{1}{c|}{--}        & \multicolumn{1}{c|}{82.9}       & \multicolumn{1}{c|}{84}         & \multicolumn{1}{c}{15.9}                             
    \\
    \multicolumn{1}{c|}{ConvNeXt-B~\cite{liu2022convnet}}  & \multicolumn{1}{c|}{--}        & \multicolumn{1}{c|}{83.8}       & \multicolumn{1}{c|}{89.0}         & \multicolumn{1}{c}{15.4}
    \\
    \multicolumn{1}{c|}{VAN-Huge~\cite{guo2022visual}}  & \multicolumn{1}{c|}{--}        & \multicolumn{1}{c|}{84.2}       & \multicolumn{1}{c|}{60.3}         & \multicolumn{1}{c}{12.2}
    \\
    \multicolumn{1}{c|}{EfficientNetV2-M~\cite{tan2021efficientnetv2}}  & \multicolumn{1}{c|}{--}        & \multicolumn{1}{c|}{85.1}       & \multicolumn{1}{c|}{54}         & \multicolumn{1}{c}{24.0}                             
    \\ 
    \multicolumn{1}{c|}{EfficientNetV2-L~\cite{tan2021efficientnetv2}}  & \multicolumn{1}{c|}{--}       & \multicolumn{1}{c|}{85.7}       & \multicolumn{1}{c|}{120}         & \multicolumn{1}{c}{53.0}                            
    \\ 
    \multicolumn{1}{c|}{PolyLoss(EfficientNetV2-L)~\cite{leng2021polyloss}}  & \multicolumn{1}{c|}{--}       & \multicolumn{1}{c|}{\textbf{87.2}}       & \multicolumn{1}{c|}{--}         & \multicolumn{1}{c}{--}             
    \\ \rowcolor{LightCyan}
    \multicolumn{1}{c|}{EfficientNetV2-XL~\cite{tan2021efficientnetv2}}  & \multicolumn{1}{c|}{ImageNet-21k}       & \multicolumn{1}{c|}{87.3}       & \multicolumn{1}{c|}{208}         & \multicolumn{1}{c}{94.0}  
    \\ \rowcolor{LightCyan}
    \multicolumn{1}{c|}{RepLKNet-XL~\cite{ding2022RepLKNet}}  & \multicolumn{1}{c|}{ImageNet-21k}       & \multicolumn{1}{c|}{\textbf{87.8}}       & \multicolumn{1}{c|}{335}         & \multicolumn{1}{c}{128.7}
    \\ \rowcolor{Gray}
    \multicolumn{1}{c|}{Meta Pseudo Labels(EfficientNet-L2)~\cite{pham2021meta}}  & \multicolumn{1}{c|}{JFT-300M}       & \multicolumn{1}{c|}{\textbf{90.2}}       & \multicolumn{1}{c|}{480}         & \multicolumn{1}{c}{--}                            
    \\ 
    \midrule
    \multicolumn{5}{c}{Transformer-based}                                                  \\ 
    \midrule
    \multicolumn{1}{c|}{ViT-B/16~\cite{dosovitskiy2020image}}          & \multicolumn{1}{c|}{--}    & \multicolumn{1}{c|}{77.9}       & \multicolumn{1}{c|}{86}         & \multicolumn{1}{c}{55.5}                             
    \\
    \multicolumn{1}{c|}{DeiT-B/16~\cite{touvron2021training}}         & \multicolumn{1}{c|}{--}     & \multicolumn{1}{c|}{81.8}       & \multicolumn{1}{c|}{86}         & \multicolumn{1}{c}{17.6}                            
    \\
    \multicolumn{1}{c|}{T2T-ViT-24~\cite{yuan2021tokens}}       & \multicolumn{1}{c|}{--} & \multicolumn{1}{c|}{82.3}       & \multicolumn{1}{c|}{64.1}         & \multicolumn{1}{c}{13.8}                              
    \\
    \multicolumn{1}{c|}{PVT-Large~\cite{wang2021pyramid}}       & \multicolumn{1}{c|}{--} & \multicolumn{1}{c|}{82.3}       & \multicolumn{1}{c|}{61}         & \multicolumn{1}{c}{9.8}                              
    \\
    \multicolumn{1}{c|}{Swin-B~\cite{liu2021swin}}            & \multicolumn{1}{c|}{--}    & \multicolumn{1}{c|}{83.5}       & \multicolumn{1}{c|}{88}         & \multicolumn{1}{c}{15.4}                                               
    \\
    \multicolumn{1}{c|}{Nest-B~\cite{zhang2021aggregating}}            & \multicolumn{1}{c|}{--}         & \multicolumn{1}{c|}{83.8}       & \multicolumn{1}{c|}{68}         & \multicolumn{1}{c}{17.9}                             
    \\
    \multicolumn{1}{c|}{PyramidTNT-B~\cite{han2022pyramidtnt}}             & \multicolumn{1}{c|}{--}     & \multicolumn{1}{c|}{84.1}       & \multicolumn{1}{c|}{157}         & \multicolumn{1}{c}{16.0}    
    \\
    \multicolumn{1}{c|}{CSWin-B~\cite{dong2021cswin}}          & \multicolumn{1}{c|}{--}     & \multicolumn{1}{c|}{84.2}       & \multicolumn{1}{c|}{78}         & \multicolumn{1}{c}{15.0} 
    \\
    \multicolumn{1}{c|}{CaiT-M-48-448~\cite{touvron2021going}}          & \multicolumn{1}{c|}{--}     & \multicolumn{1}{c|}{86.5}       & \multicolumn{1}{c|}{356}         & \multicolumn{1}{c}{330}                             
    \\
    \multicolumn{1}{c|}{PeCo(ViT-H)~\cite{dong2021peco}}            & \multicolumn{1}{c|}{--}         & \multicolumn{1}{c|}{\textbf{88.3}}       & \multicolumn{1}{c|}{635}         & \multicolumn{1}{c}{--}                             
    \\ \rowcolor{LightCyan}
    \multicolumn{1}{c|}{ViT-L/16~\cite{dosovitskiy2020image}}          & \multicolumn{1}{c|}{ImageNet-21k}    & \multicolumn{1}{c|}{85.3}       & \multicolumn{1}{c|}{307}         & \multicolumn{1}{c}{--}                             
    \\ \rowcolor{LightCyan}
    \multicolumn{1}{c|}{SwinV1-L~\cite{liu2021swin}}          & \multicolumn{1}{c|}{ImageNet-21k}    & \multicolumn{1}{c|}{87.3}       & \multicolumn{1}{c|}{197}         & \multicolumn{1}{c}{103.9}     
    \\ \rowcolor{LightCyan}
    \multicolumn{1}{c|}{SwinV2-G~\cite{liu2021swinv2}}          & \multicolumn{1}{c|}{ImageNet-21k}    & \multicolumn{1}{c|}{\textbf{90.2}}       & \multicolumn{1}{c|}{3000}         & \multicolumn{1}{c}{--}                             
    \\ \rowcolor{Gray}
    \multicolumn{1}{c|}{V-MoE~\cite{riquelme2021scaling}}          & \multicolumn{1}{c|}{JFT-300M}    & \multicolumn{1}{c|}{90.4}       & \multicolumn{1}{c|}{14700}         & \multicolumn{1}{c}{--}                         
    \\ \rowcolor{Gray}
    \multicolumn{1}{c|}{ViT-G/14~\cite{dosovitskiy2020image}}          & \multicolumn{1}{c|}{JFT-300M}    & \multicolumn{1}{c|}{\textbf{90.5}}       & \multicolumn{1}{c|}{1843}         & \multicolumn{1}{c}{--}                             
    \\
    \midrule
    \multicolumn{5}{c}{CNN + Transformer}
    \\ \midrule
    \multicolumn{1}{c|}{Twins-SVT-B~\cite{chu2021twins}}            & \multicolumn{1}{c|}{--}         & \multicolumn{1}{c|}{83.2}       & \multicolumn{1}{c|}{56}         & \multicolumn{1}{c}{8.6}                             
    \\
    \multicolumn{1}{c|}{Shuffle-B~\cite{huang2021shuffle}}            & \multicolumn{1}{c|}{--}         & \multicolumn{1}{c|}{84.0}       & \multicolumn{1}{c|}{88}         & \multicolumn{1}{c}{15.6}                             
    \\
    \multicolumn{1}{c|}{CMT-B~\cite{guo2021cmt}}            & \multicolumn{1}{c|}{--}         & \multicolumn{1}{c|}{84.5}       & \multicolumn{1}{c|}{45.7}         & \multicolumn{1}{c}{9.3}                             
    \\
    \multicolumn{1}{c|}{CoAtNet-3~\cite{dai2021coatnet}}            & \multicolumn{1}{c|}{--}         & \multicolumn{1}{c|}{84.5}       & \multicolumn{1}{c|}{168}         & \multicolumn{1}{c}{34.7}                        
    \\
    \multicolumn{1}{c|}{VOLO-D3~\cite{yuan2021volo}}            & \multicolumn{1}{c|}{--}         & \multicolumn{1}{c|}{85.4}       & \multicolumn{1}{c|}{86}         & \multicolumn{1}{c}{20.6}                             
    \\
    \multicolumn{1}{c|}{VOLO-D5~\cite{yuan2021volo}}            & \multicolumn{1}{c|}{--}         & \multicolumn{1}{c|}{\textbf{87.1}}       & \multicolumn{1}{c|}{296}         & \multicolumn{1}{c}{69.0}                             
    \\ \rowcolor{LightCyan}
    \multicolumn{1}{c|}{CoAtNet-4~\cite{dai2021coatnet}}            & \multicolumn{1}{c|}{ImageNet-21k}         & \multicolumn{1}{c|}{\textbf{88.1}}       & \multicolumn{1}{c|}{275}         & \multicolumn{1}{c}{360.9}                        
    \\ \rowcolor{Gray}
    \multicolumn{1}{c|}{CoAtNet-7~\cite{dai2021coatnet}}            & \multicolumn{1}{c|}{JFT-300M}         & \multicolumn{1}{c|}{\textbf{90.9}}       & \multicolumn{1}{c|}{2440}         & \multicolumn{1}{c}{--}                             
    \\
    \midrule
    \multicolumn{5}{c}{MLP-based}                                                          
    \\ \midrule
    \multicolumn{1}{c|}{DynaMixer-L~\cite{wang2022dynamixer}}         & \multicolumn{1}{c|}{--}    & \multicolumn{1}{c|}{\textbf{84.3}}       & \multicolumn{1}{c|}{97}       & \multicolumn{1}{c}{27.4} 
    \\
    \rowcolor{LightCyan}
    \multicolumn{1}{c|}{ResMLP-B24/8~\cite{touvron2021resmlp}}         & \multicolumn{1}{c|}{ImageNet-21k} & \multicolumn{1}{c|}{\textbf{84.4}}       & \multicolumn{1}{c|}{129.1}         & \multicolumn{1}{c}{100.2}                       
    \\ 
    \rowcolor{Gray}
    \multicolumn{1}{c|}{Mixer-H/14~\cite{tolstikhin2021mlp}}        & \multicolumn{1}{c|}{JFT-300M} & \multicolumn{1}{c|}{\textbf{86.3}}       & \multicolumn{1}{c|}{431}         & \multicolumn{1}{c}{--}
    \\
    \bottomrule
    \end{tabular}
    }
    \label{table:ImageNet}
\end{table}

From Table~\ref{table:ImageNet-MLP} and Table~\ref{table:ImageNet}, it is concluded that: (1) MLP-like models can achieve competitive performance compared to CNN-based and Transformer-based architectures with the same training strategy and data volume, (2) the performance gains brought by increasing data volume and architecture-specific training strategies may be greater than the module redesign, (3) the visual community is encouraged to build self-supervised methods and appropriate training strategies for pure MLPs.

\subsection{Object Detection and Semantic Segmentation}

Some MLP-like variants~\cite{tang2021image,zhang2021morphmlp,lian2021mlp,chen2021cyclemlp,wei2022activemlp,guo2021hire,zheng2022mixing} pre-trained on ImageNet are transferred to downstream tasks such as object detection and semantic segmentation. Such tasks are more challenging than classification due to involving multiple objects of interest in one input image. However, we currently do not have a pure MLP framework for object detection and segmentation. These MLP variants are used as backbone networks to traditional CNN-based frameworks, such as Mask R-CNN~\cite{he2017mask} and UperNet~\cite{xiao2018unified}, requiring the variant to have a pyramidal structure and resolution insensitivity.

\begin{table}[]
\renewcommand\arraystretch{0.65}
    \centering
    \caption{
       Object detection and instance segmentation results of different backbones on the COCO val2017 dataset~\cite{lin2014microsoft}. Employing the Mask R-CNN~\cite{he2017mask}, where "1x" means that a single-scale training schedule is used.
    }
    \begin{tabular}{c|cccccccc}
        \toprule
        \multirow{2}{*}{Backbone} & \multicolumn{8}{c}{Mask R-CNN 1×~\cite{he2017mask}}                                                                         \\ \cline{2-9} 
                                  & $\text{AP}^{b}$  & $\text{AP}^{b}_{50}$ & \multicolumn{1}{c|}{$\text{AP}^{b}_{75}$} & $\text{AP}^{m}$  & $\text{AP}^{m}_{50}$ & \multicolumn{1}{c|}{$\text{AP}^{m}_{75}$} & Params & FLOPs \\ 
        \midrule
        \multicolumn{9}{c}{CNN-based}                                                          
        \\ \midrule
        ResNet101~\cite{he2016deep} & 40.4 & 61.1 & \multicolumn{1}{c|}{44.2} & 36.4 & 57.7 & \multicolumn{1}{c|}{38.8}  & 63.2M  & 336G  \\
        ResNeXt101~\cite{xie2017aggregated} & 42.8 & 63.8  & \multicolumn{1}{c|}{47.3} & 38.4 & 60.6 & \multicolumn{1}{c|}{41.3}  & 101.9M & 493G  \\
        VAN-Large~\cite{guo2022visual} & 47.1 & 67.9  & \multicolumn{1}{c|}{51.9}  & 42.2 & 65.4  & \multicolumn{1}{c|}{45.5}  & 64.4M & --  \\
        \midrule
        \multicolumn{9}{c}{Transformer-based}  
        \\ \midrule
        PVT-Large~\cite{wang2021pyramid} & 42.9 & 65.0  & \multicolumn{1}{c|}{46.6}  & 39.5  & 61.9  & \multicolumn{1}{c|}{42.5}  & 81M  & 364G  \\
        Swin-B~\cite{liu2021swin} & 46.9 & --  & \multicolumn{1}{c|}{--}  & 42.3 & --  & \multicolumn{1}{c|}{--}  & 107M & 496G  \\
        CSWin-B~\cite{dong2021cswin} & \textbf{48.7} & \textbf{70.4}  & \multicolumn{1}{c|}{\textbf{53.9}}  & \textbf{43.9} & \textbf{67.8}  & \multicolumn{1}{c|}{47.3}  & 97M & 526G  \\
        \midrule
        \multicolumn{9}{c}{MLP-based}  
        \\ \midrule
        CycleMLP-B5~\cite{chen2021cyclemlp} & 44.1 & 65.5  & \multicolumn{1}{c|}{48.4} & 40.1 & 62.8 & \multicolumn{1}{c|}{43.0}  & 95.3M  & 421G \\
        WaveMLP-B~\cite{tang2021image} & 45.7 & 67.5  & \multicolumn{1}{c|}{50.1} & 27.8 & 49.2 & \multicolumn{1}{c|}{\textbf{59.7}}  & 75.1M  & 353G \\
        HireMLP-L~\cite{guo2021hire} & 45.9 & 67.2  & \multicolumn{1}{c|}{50.4} & 41.7 & 64.7 & \multicolumn{1}{c|}{45.3}  & 115.2M  & 443G \\
        MS-MLP-B~\cite{zheng2022mixing} & 46.4 & 67.2  & \multicolumn{1}{c|}{50.7} & 42.4 & 63.6 & \multicolumn{1}{c|}{46.4}  & 107.5M  & 557G \\
        ActiveMLP-L~\cite{wei2022activemlp} & 47.4 & 69.9  & \multicolumn{1}{c|}{52.0} & 43.2 & 67.3 & \multicolumn{1}{c|}{46.5}  & 96.0M  & -- \\
        \bottomrule
    \end{tabular}
    \label{table:COCO}
\end{table}

Table~\ref{table:COCO} reports object detection and semantic segmentation results of different backbones on the COCO val2017 dataset~\cite{lin2014microsoft}. As we limit the training strategy to Mask R-CNN 1x~\cite{he2017mask}, the results are not state-of-the-art on the COCO dataset. Table~\ref{table:ADE20K} reports semantic segmentation results of different backbones on the ADE20K~\cite{zhou2017scene} validation set, employing the Semantic FPN~\cite{kirillov2019panoptic} and UperNet~\cite{xiao2018unified} frameworks. Empirical results show that the performance of MLP-like variants on object detection and semantic segmentation is still weaker than the most advanced CNN and Transformer-based backbones.

Currently, an optimal backbone choice is Transformer-based, followed by CNN. Due to the resolution sensitivity, pure MLPs have not been used for downstream tasks. Recently, Transformer-based frameworks, e.g., DETR~\cite{carion2020end} have been proposed. Thus, we expect the proposal of a pure MLP framework. To this end, MLP still needs to be further explored in these fields.

\subsection{Low-level Vision}
Research on applying MLP to the low-level vision domains, such as image generation and processing, is just beginning. These tasks output images instead of labels or boxes, making them more challenging than high-level vision tasks such as image classification, object detection and semantic segmentation.

\begin{table}[]
\renewcommand\arraystretch{0.65}
    \centering
    \caption{
        Semantic segmentation results of different backbones on the ADE20K~\cite{zhou2017scene} validation set. Semantic FPN~\cite{kirillov2019panoptic} and UperNet~\cite{xiao2018unified} frameworks are employed.
    }
    \begin{tabular}{c|ccc|ccc}
        \toprule
        \multirow{2}{*}{Backbone} & \multicolumn{3}{c|}{Semantic FPN~\cite{kirillov2019panoptic}} & \multicolumn{3}{c}{UperNet~\cite{xiao2018unified}} \\ \cline{2-7} 
        & Params & FLOPs & mIoU (\%) & Params & FLOPs & mIoU (\%) \\ 
        \midrule
        \multicolumn{7}{c}{CNN-based}                                                          
        \\ \midrule
        ResNet101~\cite{he2016deep} & 47.5M & 260G & 38.8 & 86M & 1029G & 44.9 \\
        ResNeXt101~\cite{xie2017aggregated} & 86.4M & -- & 40.2 & -- & -- & -- \\
        VAN-Large~\cite{guo2022visual} & 49.0M & -- & 48.1 & 75M & -- & 50.1 \\
        ConvNeXt-XL~\cite{liu2022convnet} & -- & -- & -- & 391M & 3335G & 54.0 \\
        RepLKNet-XL~\cite{ding2022RepLKNet} & -- & -- & -- & 374M & 3431G & 56.0 \\
        \midrule
        \multicolumn{7}{c}{Transformer-based}                                                    \\ \midrule
        PVT-Medium~\cite{wang2021pyramid} & 48.0M & 219G & 41.6 & -- & -- & -- \\
        Swin-B~\cite{liu2021swin} & 53.2M & 274G & 45.2 & 121M & 1188G & 49.7 \\
        CSWin-B~\cite{dong2021cswin} & 81.2M & 464G & \textbf{49.9} & 109.2M & 1222G & 52.2 \\
        BEiT-L~\cite{bao2021beit} & -- & -- & -- & -- & -- & 57.0 \\
        SwinV2-G~\cite{liu2021swinv2} & -- & -- & -- & -- & -- & \textbf{59.9} \\
        \midrule
        \multicolumn{7}{c}{MLP-based}
        \\ \midrule
        MorphMLP-B~\cite{zhang2021morphmlp} & 59.3M & -- & 45.9 & -- & -- & --  \\
        CycleMLP-B5~\cite{chen2021cyclemlp} & 79.4M & 343G & 45.6 & -- & -- & --  \\
        Wave-MLP-M~\cite{tang2021image} & 43.3M & 231G & 46.8 & -- & -- & -- \\
        AS-MLP-B~\cite{lian2021mlp} & -- & -- & -- & 121M & 1166G & 49.5 \\ 
        HireMLP-L~\cite{guo2021hire} & -- & -- & -- & 127M & 1125G & 49.9 \\ 
        MS-MLP-B~\cite{zheng2022mixing} & -- & -- & -- & 122M & 1172G & 49.9 \\ 
        ActiveMLP-L~\cite{wei2022activemlp} & 79.8M & -- & 48.1 & 108M & 1106G & 51.1 \\
        \bottomrule
    \end{tabular}

    \label{table:ADE20K}
\end{table}

Cazenavette \emph{et al.}~\cite{cazenavette2021mixergan} propose MixerGAN for unpaired image-to-image translation. Specially, MixerGAN adopts the framework of CycleGAN~\cite{zhu2017unpaired}, but replaces the convolution-based Residual Block with the Mixer Layer of MLP-Mixer. Their experiments show that the MLP-Mixer succeeds at generative objectives, and although being an initial exploration, it is promising in extending the MLP-based architecture to image composition tasks further.

Tu \emph{et al.}~\cite{tu2022maxim} propose MAXIM, a UNet-shaped hierarchical structure that supports long-range interactions enabled by spatially-gated MLPs. MAXIM contains two MLP-based building blocks: a multi-axis gated MLP and a cross-gating block, both are variants of the gMLP block~\cite{liu2021pay}. By applying gMLP to low-level vision tasks to gain global information, the MAXIM family has achieved state-of-the-art performance in multiple image processing tasks with moderate complexity, including image dehazing, deblurring, denoising, deraining, and enhancement.

\subsection{Video Analysis}
Several works extend MLP to temporal modeling and video analysis. MorphMLP~\cite{zhang2021morphmlp} gets competitive performance with recent state-of-the-art methods on Kinetics-400~\cite{carreira2017quo} dataset, demonstrating that MLP-like backbone is also suitable for video recognition. Skating-Mixer~\cite{xia2022skating} extends the MLP-Mixer-based framework to learn from video. It is used to figure skating scoring at the Beijing 2022 Winter Olympic Games and demonstrates satisfactory performance. However, compared with other methods, the number of frames in a single input has not been increased. Therefore, their advantage may be the larger spatial receptive field, instead of capturing long-term temporal information.

\subsection{Point Cloud}

Point cloud analysis can be considered a special vision task, which is increasingly used in real-time by robots and self-driving vehicles to understand their environment and navigate through it. Unlike images, point clouds are inherently sparse, unordered and irregular. The unordered nature is one of the biggest challenges for CNNs based on local receptive fields, because input adjacent does not imply spatial adjacent. In contrast, MLP is naturally invariant to permutation, which perfectly fits the characteristic of point cloud~\cite{ma2022rethinking}, making classical frameworks like PointNet~\cite{qi2017pointnet} and PointNet++~\cite{qi2017pointnet++} MLP-based.

Choe \emph{et al}.~\cite{choe2021pointmixer} design PointMixer, which embeds geometric relations between points features into the MLP-Mixer’s framework. The relative position vector is utilized for processing unstructured 3D points, and Token-mixing MLP is replaced with a softmax function. Ma \emph{et al}.~\cite{ma2022rethinking} construct a pure residual MLP network, called PointMLP. It introduces a lightweight geometric affine module to transform the local points to a normal distribution. It then employs simple residual MLPs to represent local points, as they are permutation-invariant and straightforward. PointMLP achieves the new state-of-the-art performance on multiple datasets. Additionally, recently proposed Transformer-based networks~\cite{guo2021pct,zhao2021point} show competitive performance, where self-attention is permutation invariant for processing a sequence of points, making it well-suited for point cloud learning.

From the above analysis, it is evident that Transformer and MLP are appealing solutions for unordered data, where disorder makes it challenging to design artificial inductive biases.

\subsection{Discussion}
MLP-like variants have been applied for diverse vision tasks, such as image classification, image generation, object detection, semantic segmentation, and video analysis, achieving outstanding performance due to the artificial redesigning of the MLP block. Nevertheless, constructing MLP frameworks and employing MLP-specific training strategies may improve performance further. Additionally, pure MLPs have already demonstrated their advantages in point cloud analysis, encouraging the application of MLPs to visual tasks with unordered data.

\section{Summary and Outlook}
\label{sec:Outlook}

As the history of computer vision attests, the availability of larger datasets along with the increase in computational power often triggers a paradigm shift. And within these paradigm shifts, there is a gradual reduction in human intervention, i.e., removing hand-crafted inductive biases and allowing the model to further freely learn from the raw data~\cite{tolstikhin2021mlp}. The MLP and Boltzmann machines proposed in the last century exceeded the computational conditions at the time and were not widely used. In contrast, computationally efficient CNNs are more popular and replace manual feature extraction. From CNNs to Transformer, we have seen the models' receptive field expand step by step, and the spatial range considered when encoding features is getting larger and larger. From Transformer to deep MLP, we no longer use similarity as the weight matrix, but let the model learn the weights from the raw data. The latest MLP works all seem to suggest that \emph{Deep MLPs are making a strong comeback as the new paradigm}. In the latest MLP development, we see compromises such as:
\begin{enumerate}[(1)]
\item The latest proposed deep MLP-based models use patch partition instead of flattening the entire input to compromise computational cost. This allows the full connectivity and global receptive field to be approximated at the patch level. The patch partition forms a two-dimensional matrix $hw \times Cp^2$, instead of a one-dimensional vector $1 \times HWC$ as in the entire input flattening case, where $H = ph, W = pw$ are the input resolution, $p$ is the patch size, and $C$ is the input channel. Subsequently, the fully connected projections are performed alternately on space and channels. This is an orthogonal decomposition of traditional fully connected projections, just as the full space projection is further orthogonalized into horizontal and vertical directions.
\item At the module design level, there are two main improvement routes. One MLP-like variant type focuses on reducing computational complexity, which compromises computational power, and this reduction comes at the expense of decoupling the full spatial connectivity. Another type of variant addresses the resolution-sensitivity problem, making it possible to transfer pre-trained models to downstream tasks. These works adopt CNN-like improvements, but the full connectivity and global receptive field as in MLPs are eroded. The receptive field evolves in the opposite direction in these models, becoming smaller and smaller and backing to the CNN ways. 
\item At the architecture level, the traditional block stacking patterns are also applicable to MLP, and it seems that the pyramid structure is still the best choice, with the initial smaller patch size helping to obtain finer features. Note that this comparison is unfair because the initial patch of the current single-stage model is larger ($16\times 16$), and the initial patch of the pyramid model is smaller ($4\times 4$). The pyramid structure compromises small patches and low computational costs to some extent. What if a $4\times 4$ patch partition is used in a single-stage model? Will it be worse than the pyramid model? This is unknown. What is known is that the cost of calculations will increase significantly, and it strains the current computing devices.
\end{enumerate}

The results of our research suggest that the current amount of data and computational capability are still not enough to support pure vision MLP models to learn effectively. Moreover, human intervention still occupies an important place. Based on this conclusion, we elaborate on potential future research directions.

\textbf{Vision Tailored Designs}. With the current amount of data and computation, human guidance remains important, and it seems natural to combine the advantages of other architectures~\cite{liu2021uninet,valanarasu2022unext}. Currently, most MLP-like variants remain an either/or choice for short-range and long-range dependencies and need further intuitions to enhance their efficiency on visual inputs. RepMLPNet~\cite{ding2021repmlpnet} has made a viable attempt. We believe that in the future, the community should focus on how to combine short-range dependencies and long-range dependencies rather than keeping only one or the other. This is consistent with human intuition as local details are beneficial for understanding individual objects and the interactions across the entire visual field remain significant. Note that image resolution insensitivity is important, ensuring the network to be a universal vision backbone. To sum up, we encourage the community to rethink tailored visual designs further, i.e., to integrate the global receptive field (long-range dependencies) and local receptive field (short-range dependencies) while maintaining resolution insensitivity.

\textbf{Scaling-UP/Down Techniques}. It has long been recognized that larger vision models generally perform better on vision tasks~\cite{he2016deep,simonyan2014very}, but the size of most vision networks is only a few million to over 100 million. Furthermore, various configurations of an MLP-like variant offer limited gains despite the increased number of parameters. Recently, the visual community has conducted some scaling-up research on the vision Transformer with self-supervised pre-training, including V-MoE~\cite{riquelme2021scaling}, Swinv2~\cite{liu2021swinv2} and ViTAEv2~\cite{zhang2022vitaev2}, which afford a considerable performance boost. Nevertheless, scaling-up techniques specific to MLPs need further exploration. Moving from the lab to life, MLPs can have intensive power and computation requirements, hindering their deployment on edge devices and resource-constrained environments such as mobile phone platforms. Such hardware efficient designs are currently lacking for the vision MLPs to enable their seamless deployment in resource-constrained devices. How do MLPs perform with low precision training and inference? How do MLPs perform knowledge distillation? How about using Neural Architecture Search~\cite{liu2018progressive} to design more efficient and light-weight MLP models? It will be interesting to see how these questions are answered.

\textbf{Dedicated Pre-Training and Optimizing Method}. MLPs enjoy greater freedom and larger solution space with less inductive biases than CNNs and Transformer. We currently have difficulties finding optimal solutions for MLPs, which is commonly attributed to computation power and data volume constraints. Pre-training helps generalization and the very limited information in the labels is used only to slightly adjust the weights found by pre-training. Many self-supervised frameworks, such as MoCo~\cite{he2020momentum,chen2020improved,chen2021empirical}, SimCLR~\cite{chen2020simple,chen2020big}, SimMIM~\cite{xie2021simmim}, MAE~\cite{he2021masked}, MaskFeat~\cite{wei2021masked}, have already provided a big boost to CNNs and Transformer. Are these self-supervised learning methods still effective for MLPs? Can a better self-supervised training method be designed for MLPs? What's more, is this related to the optimizer used? We know that SGD~\cite{bottou2012stochastic} is a good optimizer for CNNs and AdamW~\cite{loshchilov2018fixing} performs well for Transformer. What is the best choice for MLPs? A recent work~\cite{chen2021vision} has conducted a preliminary exploration and investigates the MLP-Mixer from the lens of loss landscape geometry. GGA-MLP~\cite{bansal2022gga} proposes a greedy genetic algorithm to optimize weights and biases in MLP. We believe that dedicated pre-training and optimizing methods will be an excellent boost to accelerate the development of deep MLP models.

\textbf{Interpretability}. Another optional direction is towards a more in-depth analysis and comparison of the filters learned by the network and the resulting feature maps. MLPs continue the long-term trend of removing hand-crafted inductive biases and allowing models to further freely learn from the raw data. What follows is that the interpretability of the model is getting lower and lower. Both mathematical explanations and visual analysis are possible helpful to understand what neural networks can freely learn from massive amounts of raw data with fewer priors. This can assist in proving whether some of the past artificial priors are correct or incorrect and potentially guide the design choices of future networks. In addition, the theoretical understanding of why networks might be vulnerable is also a key topic.

\textbf{Beyond MLP}. With the further improvement of the data volume and computing power, we should be beyond the horizon of present knowledge, the weighted-sum paradigms, and reconsider more theoretical paradigms from mathematic and physic systems, such as Boltzmann machines. The weighted-sum paradigms have driven the booming of GPU-based computing and deep learning itself, while we believe the Boltzmann-like paradigm will also grow up with a new generation of computing hardware.

\section*{Acknowledgments}
This research is supported by the National Key Research and Development Program of China (Grant No.2021ZD0113801 and 2021YFC3330202), Beijing Academy of Artificial Intelligence (BAAI), and Basic Research Fund of Shenzhen City (Grant No.JCYJ20210324120012033). We would like to thank Prof. Shi-min Hu and Dr. Zhaowei Xi for their helpful discussions and insightful suggestions. 

We acknowledge and deeply appreciate all the feedback and comments provided by the editors and the panel of anonymous reviewers. Your work has greatly enhanced the quality and contribution of this article.

\section*{Declaration of Interests}
The authors declare no competing interests.

\section*{Data and Code Availability}
We reproduce most variants of MLP-like models in Jittor and Pytorch. Code is available at \url{https://github.com/liuruiyang98/Jittor-MLP}. This review did not generate new data.

\section*{Author Contributions}
L.T. designed the roadmap of the article. L.T. and R.L. initiated the project and supervised the team in collaboration with Y.L. D.L. H.Z., and L.T. R.L. Y.L. drafted the manuscript. R.L. and D.L. reproduced and open-sourced the code of MLP-like variants in Jittor. All co-authors participated in discussions, and revised the manuscript iteratively.

\bibliography{mybibfile}

\end{document}